\documentclass[9pt,twocolumn,twoside]{pnas-new}

\templatetype{pnasresearcharticle} 

\usepackage{nameref}
\usepackage{url}
\usepackage{graphicx}
\usepackage{subfig}
\usepackage{url}

\usepackage{array}
\usepackage{multirow}
\usepackage{cellspace}
\usepackage{float}

\title{Automatically identifying, counting, and describing wild animals in camera-trap images with deep learning}

\author[1]{Mohammed Sadegh Norouzzadeh}
\author[2]{Anh Nguyen} 
\author[3]{Margaret Kosmala} 
\author[4]{Ali Swanson} 
\author[5]{Meredith Palmer} 
\author[5]{Craig Packer} 
\author[1,6]{Jeff Clune}

\affil[1]{University of Wyoming}
\affil[2]{Auburn University}
\affil[3]{Harvard University}
\affil[4]{University of Oxford}
\affil[5]{University of Minnesota}
\affil[6]{Uber AI Labs}

\leadauthor{Norouzzadeh}

\correspondingauthor{\textsuperscript{2}To whom correspondence should be addressed. E-mail: jeffclune@uwyo.edu}


\begin{abstract}
Having accurate, detailed, and up-to-date information about the location and behavior of animals in the wild would revolutionize our ability to study and conserve ecosystems. We investigate the ability to automatically, accurately, and
inexpensively collect such data, which could transform many fields of biology,
ecology, and zoology into ``big data'' sciences. Motion sensor ``camera traps'' enable collecting wildlife pictures inexpensively,
unobtrusively, and frequently. However, extracting information from these pictures remains an expensive, time-consuming, manual task. We demonstrate that such information can be
automatically extracted by deep learning,
a cutting-edge type of artificial intelligence. We train deep convolutional neural networks to identify, count, and
describe the behaviors of 48 species in the 3.2-million-image Snapshot Serengeti dataset. Our deep neural networks automatically identify animals with over
93.8\% accuracy, and we expect that number to improve rapidly in years to come.
More importantly, if our system classifies only images it
is confident about, our system can automate animal identification
for 99.3\% of the data while still performing at the same 96.6\% accuracy as that
of crowdsourced teams of human volunteers, saving more than ~8.4 years (at 40
hours per week) of human labeling effort (i.e. over 17,000 hours) on this
3.2-million-image dataset. Those efficiency gains immediately highlight the
importance of using deep neural networks to automate data extraction from
camera-trap images. Our results suggest that this technology could enable the
inexpensive, unobtrusive, high-volume, and even real-time collection of a
wealth of information about vast numbers of animals in the wild.
	
\end{abstract}

\dates{Last edited on \today}
\doi{\url{www.pnas.org/cgi/doi/10.1073/pnas.XXXXXXXXXX}}
\begin{document}

\verticaladjustment{-2pt}

\maketitle
\thispagestyle{firststyle}
\ifthenelse{\boolean{shortarticle}}{\ifthenelse{\boolean{singlecolumn}}{\abscontentformatted}{\abscontent}}{}

\begin{figure}[h!]
	\setlength{\unitlength}{0.14in}
	\centering 
	\includegraphics[width=0.5\textwidth]{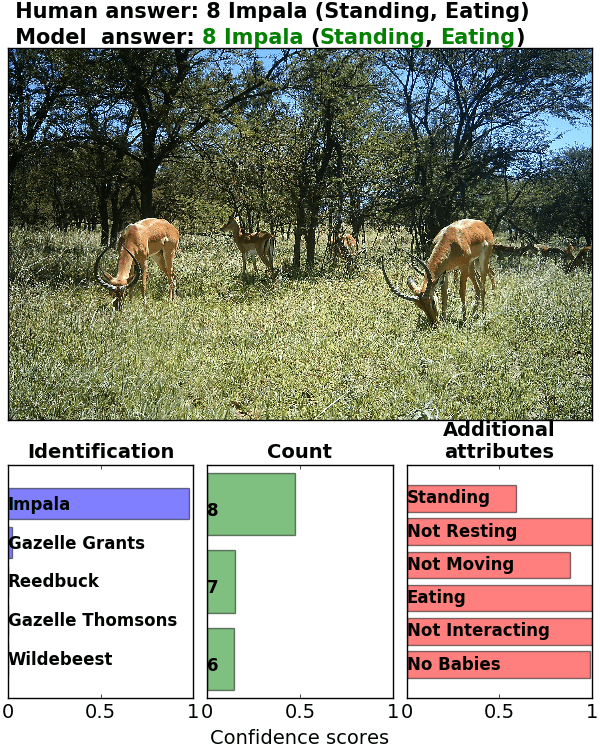}
	\caption{Deep neural networks can successfully identify, count, and describe animals in camera-trap images. Above the image: the ground-truth, human-provided answer (top line) and the prediction (second line) by a deep neural network we trained (ResNet-152). The three plots below the image, from left to right, show the neural network's prediction for the species,  number, and behavior of the animals in the image. The horizontal color bars indicate how confident the neural network is about its predictions. All similar images in this paper are from the Snapshot Serengeti dataset \cite{swanson2015snapshot}.}
	\label{fig:example}
\end{figure}

\begin{figure*}[h!]
	\centering
	\subfloat[Partially visible animal (left)]{{\includegraphics[width=0.24\textwidth]{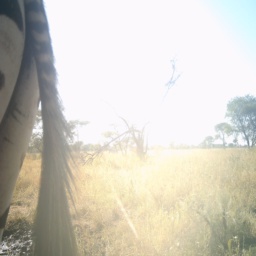} }}%
	\hskip 1pt
	\subfloat[Far away animals (center)]{{\includegraphics[width=0.24\textwidth]{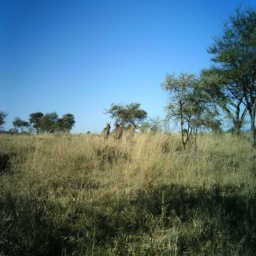} }}%
	\hskip 1pt
	\subfloat[Close-up shot of an animal]{{\includegraphics[width=0.24\textwidth]{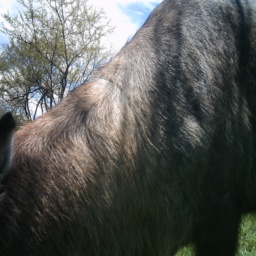} }}%
	\hskip 1pt
	\subfloat[Image taken at night]{{\includegraphics[width=0.24\textwidth]{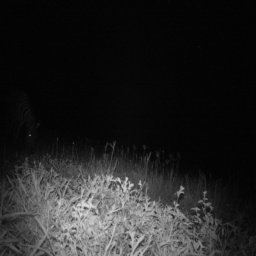} }}%
	
	\caption{Various factors make identifying animals in the wild hard even for humans (trained volunteers achieve 96.6\% accuracy vs. experts).}
	\label{fig:IdentificationDifficulties}
\end{figure*}

\dropcap{T}o better understand the complexities of natural ecosystems and better manage and protect them, it would be helpful to have detailed, large-scale knowledge about the number, location, and behaviors of animal in natural ecosystems \cite{harris2010automatic}. 
Placing motion sensor cameras called ``camera traps'' in natural habitats has revolutionized wildlife ecology and conservation over the last two decades \cite{o2010camera}. These camera traps have become an essential tool for ecologists, enabling them to study population sizes and distributions \cite{silveira2003camera}, and evaluate habitat use \cite{bowkett2008use}. While they can take millions of images \cite{fegraus2011data,krishnappa2014software,swinnen2014novel}, extracting knowledge from these camera-trap images is traditionally done by humans (i.e. experts or a community of volunteers) and is so time-consuming and costly that much of the invaluable knowledge in these big data repositories remains untapped. For example, currently it takes 2-3 months for thousands of ``citizen scientists''  \cite{swanson2015snapshot} to label each 6-month batch of images for the Snapshot Serengeti (hereafter, SS). By 2011, there were 125 camera-trap projects worldwide \cite{fegraus2011data}, and, as digital cameras become better and cheaper, more projects will put camera traps into action. Most of these projects, however, are not able to recruit and harness a huge volunteer force as SS has done. In other words, most of the invaluable information contained in raw camera-trap images may be wasted. Automating the information extraction procedure (Fig. \ref{fig:example}) will thus make vast amounts of invaluable information easily available for ecologists to help them perform their scientific, management, and protection missions.

In this paper, we focus on harnessing computer vision to automatically extract the species, number, presence of young, and behavior (e.g. moving, resting, or eating) of animals. These tasks can be challenging even for humans. Images taken from camera traps are rarely perfect, and many images contain animals that are far away, too close, or only partially visible (Fig. \ref{fig:IdentificationDifficulties}a-c). In addition, different lighting conditions, shadows, and weather can make the information extraction task even harder (Fig. \ref{fig:IdentificationDifficulties}d). 
Human-volunteer species and count labels are estimated to be 96.6\% and 90.0\% accurate, respectively, vs. labels provided by experts \cite{swanson2015snapshot}.

Automatic animal identification and counting would improve all biology missions that require identifying species and counting individuals, including animal monitoring and management, examining biodiversity, and population estimation \cite{o2010camera}.
In this paper, we harness deep learning, a state-of-the-art machine learning technology that has led to dramatic improvements in artificial intelligence in recent years, especially in computer vision \cite{Goodfellow-et-al-2016-Book}. 
 
Deep learning only works well with vast amounts of labeled data, significant computational resources, and modern neural network architectures. Here, we combine the millions of labeled data from the SS project, modern supercomputing, and state-of-the-art deep neural network (DNN) architectures to test whether deep learning can automate information extraction from camera-trap images. We find that the system is both able to perform as well as teams of human volunteers on a large fraction of the data, and identifies the few images that require human evaluation. The net result is a system that dramatically improves our ability to automatically extract valuable knowledge from camera-trap images.

\section*{Background and Related Work}

\subsection*{Machine Learning}
\label{sec:machineLearning}
Machine learning enables computers to solve tasks without being explicitly programmed to solve them \cite{samuel1959some}. State-of-the-art methods teach machines via \emph{supervised learning} i.e. by showing them correct pairs of inputs and outputs \cite{mohri2012foundations}. For example, when classifying images, the machine is trained with many pairs of images and their corresponding labels, where the image is the input and its correct label (e.g. ``Buffalo'') is the output (Fig.~\ref{fig:DeepLearning}). 

\subsection*{Deep Learning}
\begin{figure}[!h]
	\centering 
	\includegraphics[width=0.5\textwidth]{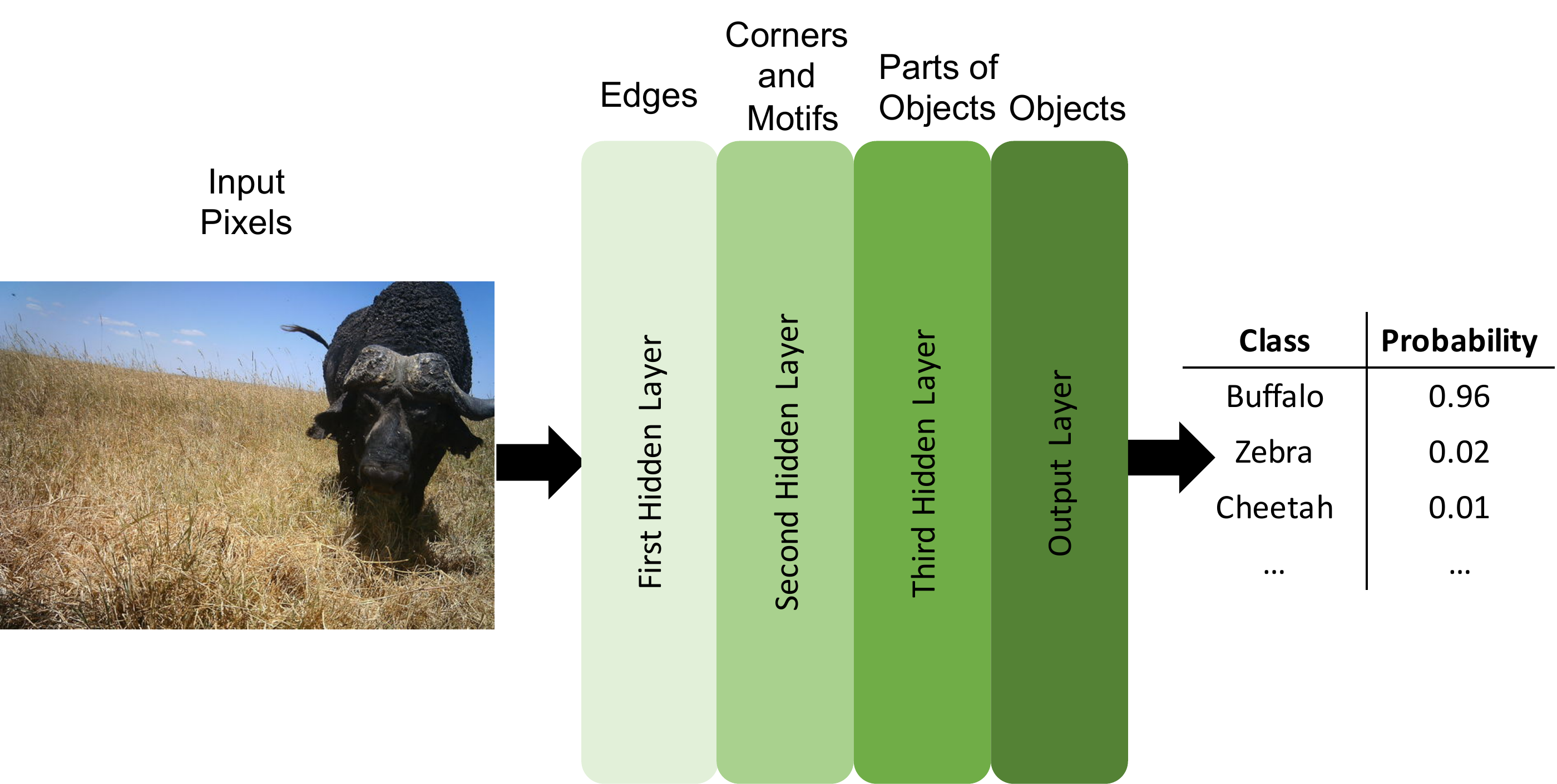}
	\caption{Deep neural networks have several layers of abstraction that tend to gradually convert raw data into more abstract concepts. For example, raw pixels at the input layer are first processed to detect edges (first hidden layer), then corners and textures (second hidden layer), then object parts (third hidden layer), and so on if there are more layers, until a final prediction is made by the output layer. Note that which types of features are learned at each layer are not human-specified, but emerge automatically as the network learns how to solve a given task.
	}
	\label{fig:DeepLearning} 
\end{figure}

Deep learning \cite{lecun2015deep} allows the machine to automatically extract multiple levels of abstraction from raw data (Fig. \ref{fig:DeepLearning}). 
Inspired by the mammalian visual cortex \cite{hu2015deep}, deep convolutional neural networks are a class of feedforward DNNs\cite{lecun2015deep} in which each layer of neurons employs convolutional operations to extract information from overlapping small regions coming from the previous layers \cite{Goodfellow-et-al-2016-Book}. The final layer of a DNN is usually a softmax function, with an output between 0 and 1 per class, and with all of the class outputs summing to 1. These outputs are often interpreted as the DNN's estimated probability of the image belonging in a certain class, and higher probabilities are often interpreted as the DNN being more confident that the image is of that class \cite{bridle1990probabilistic}. DNNs have dramatically improved the state of the art in many challenging problems \cite{Goodfellow-et-al-2016-Book}, including speech recognition \cite{hinton2012deep,deng2013new,bahdanau2016end}, machine translation  \cite{sutskever2014sequence,cho2014learning}, image recognition \cite{he2015deep,simonyan2014very}, and playing Atari games \cite{mnih2015human}. 

\subsection*{Related Work} 

There have been many attempts to automatically identify animals in camera-trap images; however, many relied on hand-designed features \cite{swinnen2014novel,figueroa2014fast, wang2014automatic} to detect animals, or were applied to small datasets (e.g. only a few thousand images) \cite{yu2013automated,chen2014deep,wang2014automatic}. In contrast, in this work, we seek to (a) harness deep learning to automatically extract necessary features to detect, count, and describe animals; and (b) apply our method on the world's largest dataset of wild animals i.e. the SS dataset \cite{swanson2015snapshot}.

Previous efforts to harness hand-designed features to classify animals include Swinnen et al. \cite{swinnen2014novel}, who attempted to distinguish the camera-trap recordings that do not contain animals or the target species of interest by detecting the low-level pixel changes between frames. Yu et. al. \cite{yu2013automated} 
extracted the features with sparse coding spatial pyramid matching \cite{yang2009linear} and utilized a linear support vector machine \cite{mohri2012foundations} to classify the images. While achieving 82\% accuracy, their technique requires manual cropping of the images, which requires substantial human effort. 

Several recent works harnessed deep learning to classify camera-trap images. Chen et. al. \cite{chen2014deep} harnessed convolutional neural networks (CNNs) to fully automate animal identification. However, they demonstrated the techniques on a dataset of around 20,000 images and 20 classes, which is of much smaller scale than we explore here \cite{chen2014deep}. In addition, they obtained an accuracy of only 38\%, which leaves much room for improvement. Interestingly, Chen et al. found that DNNs outperform a traditional  Bag of Words technique \cite{blei2003latent,fei2005bayesian} if provided sufficient training data \cite{chen2014deep}. 
Similarly, Gomez et al. \cite{gomez2016animal} also had success applying DNNs to distinguishing birds vs. mammals in a small dataset of 1,572 images and distinguish two mammal sets in a dataset of 2,597 images. 

The closest work to ours is Gomez et al. \cite{gomez2016towards}, who also evaluate DNNs on the SS dataset: they perform only the species identification task, whereas we also attempt to count animals, describe their behavior, and identify the presence of young.
On the species identification task, our models perform far superior to theirs: 92.0\% for our best network vs. around 57\% (estimating from their plot, as the exact accuracy was not reported) for their best network. There are multiple other differences between our work and theirs. (a) Gomez et al. only trained networks on a simplified version of the full 48-class SS dataset. Specifically, they removed the 22 classes that have the fewest images (Fig.~\ref{fig:CaptureEvents}, bottom 22 classes) from the full dataset and thus classify only 26 classes of animals.
Here, we instead seek solutions that perform well on all 48 classes as the ultimate goal of our research is to automate as much of the labeling effort as possible. (b) Gomez et al. base their classification solutions on networks pre-trained on the ImageNet dataset \cite{deng2009imagenet}, a technique known as transfer learning \cite{yosinski2014transferable}. We found that transfer learning made very little difference on this task, and we thus chose not to use it for simplicity: see supplementary information (SI) Sec. \nameref{app:transfer}. We conduct a more detailed comparison with Gomez et al. \cite{gomez2016towards} in SI Sec.~\nameref{sec:animalIdentificationCOLO}. 

\begin{figure*}
	\centering
	\subfloat[Image 1]{{\includegraphics[width=0.3\textwidth]{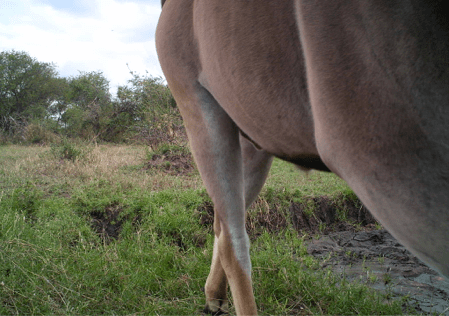} }}%
	\hskip 2pt
	\subfloat[Image 2]{{\includegraphics[width=0.3\textwidth]{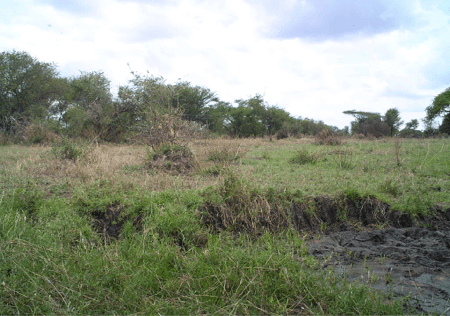} }}%
	\hskip 2pt
	\subfloat[Image 3]{{\includegraphics[width=0.3\textwidth]{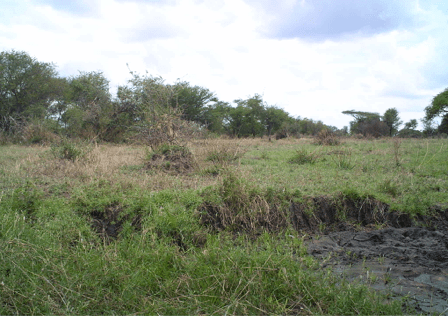} }}%
	
	\caption{While we train models on individual images, we only have labels for entire capture events, which we apply to all images in the event. When some images in an event have an animal and others are empty (as in this example), the empty images are labeled with an animal type, which introduces some noise in the training set labels and thus makes training harder.}
	\label{fig:CENoise}
\end{figure*}

\subsection*{Snapshot Serengeti Project} 
\label{sec:SS}
The Snapshot Serengeti project is the world's largest camera-trap project published to date, with 225 camera traps running continuously in Serengeti National Park, Tanzania, since 2011 \cite{swanson2015snapshot}. Whenever a camera trap is triggered, such as by the movement of a nearby animal, the camera takes a set of pictures (usually 3). Each trigger is referred to as a \emph{capture event}. The public dataset used in this paper contains 1.2 million capture events (3.2 million images) of 48 different species.

Nearly 28,000 registered and 40,000 unregistered volunteer citizen scientists have labeled 1.2 million SS capture events. For each image set, multiple users label the species, number of individuals, various behaviors (i.e. standing, resting, moving, eating, or interacting), and the presence of young. In total, 10.8 million classifications from volunteers have been recorded for the entire dataset. Swanson et al. \cite{swanson2015snapshot} developed a simple algorithm to aggregate these individual classifications into a final ``consensus'' set of labels, yielding a single classification for each image and a measure of agreement among individual answers.
In this paper, we focus on capture events that contain only one species; we thus removed events containing more than one species from the dataset (around 5\% of the events). Extending these techniques to images with multiple species is a fruitful area for future research. 
In addition to volunteer labels, for about 4,000 capture events the SS dataset also contains expert-provided labels, but only of the number and type of species present.

75\% of the capture events were classified as empty of animals. Moreover, the dataset is very unbalanced, meaning that some species are much more frequent than others (SI Sec.~\nameref{sec:Imbalance}). Such imbalance is problematic for machine learning techniques because they become heavily biased towards classes with more examples. If the model just predicts the frequent classes such as \emph{wildebeest} or \emph{zebra} most of the time, it can still get a very high accuracy without investing in learning rare classes, even though these can be of more scientific interest. The imbalance problem also exists for describing behavior and identifying the presence of young. Only 1.8\% of the capture events are labeled as containing babies; and only 0.5\% and 8.5\% of capture events are labeled as interacting and resting, respectively. We delve deeper into this problem in SI Sec.~\nameref{sec:Imbalance}. 

The volunteers labeled entire capture events (not individual images). While we do report results for labeling entire capture events (SI Sec.~\nameref{sec:capture_events}), in our main experiment, we focus on labeling individual images instead because if we ultimately can correctly label individual images it is easy to infer the labels for capture events. Importantly, we also found that utilizing individual images results in higher accuracy because it allows three times more labeled training examples (SI Sec.~\nameref{sec:capture_events}). In addition, training our system on images makes it more informative and useful for other projects, some of which are image-based and not capture-event-based.

However, the fact that we take the labels for each capture event and assign them to all the individual images in that event introduces noise into the training process. For example, a capture event may have one image with animals, but the remaining images empty (Fig. \ref{fig:CENoise}). Assigning a species label (e.g. hartebeest Fig. \ref{fig:CENoise}a) to all these images (Fig. \ref{fig:CENoise}b,c) adds some noise that machine learning models must overcome.

\section*{Experiments and Results}
\label{sec:exps}
We found that a two-stage pipeline outperforms a one-step pipeline (SI Sec.~\nameref{sec:onestep}): in the first stage a network solves the \emph{empty vs. animal} task (task I),  i.e. detecting if an image contains an animal; in the second \emph{information extraction} stage, a network then reports information about the images that contain animals. 75\% of the images are labeled empty by humans, therefore automating the first stage alone saves 75\% of human labor. 

The \emph{information extraction} stage contains three additional tasks: (II) identifying which species is present, (III) counting the number of animals, and (IV) describing additional animal attributes (their behavior and whether young are present). We chose to train one model to simultaneously perform all of these tasks, a technique ---called multitask learning \cite{caruana1998multitask}---because (a) these tasks are related, therefore they can share weights that encode features common to all tasks (e.g. recognizing animals); learning multiple, related tasks in parallel often improves the performance on each individual task \cite{collobert2008unified}, and (b) doing so requires fewer model parameters vs. a separate model for each task, meaning we can solve all tasks faster and more energy efficiently, and the model is easier to transmit and store. These advantages will become especially important if such neural network models run on remote camera traps to determine which pictures to store or transmit. 

\subsection*{Datasets}
In this paper, we only tackle identifying one instead of multiple species in an image (i.e. single-label classification \cite{mohri2012foundations}). Therefore we removed images that humans labeled as containing more than one species from our training and testing sets (approximately 5\% of the dataset). The training and test sets for the information extraction stage are formed from the 25\% of images that are labeled as non-empty by humans.

If there are overly similar images in the training and test sets, models can just memorize the examples and then do not generalize well to dissimilar images. To avoid this problem, we put entire capture events (which contain similar images) into either the training or test set. From a total of 301,400 capture events that contained an animal, we created a training set containing 284,000 capture events, and two test sets. The \emph{expert-labeled test set} contains 3,800 capture events with species and counts labels. The \emph{volunteer-labeled test set} contains 17,400 capture events labeled by volunteers and it has labels for species, counts, behaviors, and the presence of young.

\begin{table*}[h]
\caption{In this paper, we employ different deep learning architectures to infer which one works the best and to be able to compare difference between accuracies come from different architectures.}
	\centering
	\begin{tabular}{m{1.8cm}m{1.5cm}m{9cm}}
		\textbf{Architecture}&\textbf{\# of Layers}& \textbf{Short Description}\\
		\hline
		AlexNet&8&A landmark architecture for deep learning winning ILSVRC 2012 challenge \cite{krizhevsky2012imagenet}.\newline\\
		NiN&16& Network in Network (NiN) is one of the first architectures harnessing innovative 1x1 convolutions \cite{lin2013network} to provide more combinational power to the features of a convolutional layers \cite{lin2013network}.\newline\\
		VGG&22& An architecture that is deeper and obtains better performance than AlexNet by employing effective 3x3 convolutional filters \cite{simonyan2014very}.\newline
		\\
		GoogLeNet&32& This architecture is designed to be computationally efficient (using 12 times fewer parameters than AlexNet) while offering high accuracy \cite{szegedy2015going}.
		\\
		\multirow{5}{*}{\parbox{9cm}{ResNet}}&18&\multirow{5}{*}{\parbox{9cm}{The winning architecture of the 2016 ImageNet competition \cite{he2015deep}. The number of layers for the ResNet architecture can be different. In this paper, we try 18, 34, 50, 101, and 152 layers.}}\\
		&34&\\
		&50&\\
		&101&\\
		&152&\\
		\hline
	\end{tabular}
	\label{tab:layers}
\end{table*}
\subsection*{Architectures}
Different DNNs have different \emph{architectures}, meaning the type of layers they contain (e.g. convolutional layers, fully connected layers, pooling layers, etc.), and the number, order, and size of those layers \cite{Goodfellow-et-al-2016-Book}. In this paper, we test 9 different modern architectures at or near the state of the art (Table \ref{tab:layers}) to find the highest-performing networks and to compare our results to those from Gomez et al. \cite{gomez2016towards}. We only trained each model one time because doing so is computationally expensive and because both theoretical and empirical evidence suggests different DNNs trained with the same architecture, but initialized differently, often converge to similar performance levels \citep{Goodfellow-et-al-2016-Book,lecun2015deep,dauphin2014identifying}.

A well-known method for further improving classification accuracy is to employ an ensemble of models at the same time and average their predictions. After training all the nine models for each stage, we form an ensemble of the trained models by averaging their predictions (SI Sec.~\nameref{app:merge}). More details about the architectures, training methods, pre-processing steps and the hyperparameters are in Sec. \nameref{app:archs}.

\subsection{Task I: Detecting Images That Contain Animals}
\label{sec:task1}
For this task, our models take an image as input and output two probabilities describing whether the image has an animal or not (i.e. binary classification). We train 9 neural network models (Table~\ref{tab:layers}). 
Because 75\% of the SS dataset is labeled as empty, to avoid imbalance between the empty and non-empty classes, we take all 25\% (757,000) non-empty images and randomly select 757,000 ``empty'' images. This dataset is then split it into training and test sets. 

The training set contains 1.4 million images and the test set contains 105,000 images. Since the SS dataset contains labels for only capture events (not individual images), we assign the label of each capture event to all of the images in that event. All the architectures achieve a classification accuracy of over 95.8\% on this task. The VGG model achieved the best accuracy of 96.8\% (Table~\ref{tab:EF}). 
To show the difficulty of the task and where the models currently fail, several examples for the best model (VGG) are shown in SI Sec.~\nameref{sec:silvertestset}.

\begin{table}
	\caption{Accuracy of different models on \nameref{sec:task1}}
	\centering
	\setlength{\unitlength}{0.10in}
	\begin{tabular}{p{2.5cm}p{1.9cm}}
		\textbf{Architecture}&\textbf{Top-1 accuracy}\\
		\hline
		AlexNet&95.8\%\\
		NiN&96.0\%\\
		\textbf{VGG}&\textbf{96.8}\%\\
		GoogLeNet&96.3\%\\
		ResNet-18&96.3\%\\
		ResNet-34&96.2\%\\
		ResNet-50&96.3\%\\
		ResNet-101&96.1\%\\
		ResNet-152&96.1\%\\
		Ensemble of models&96.6\%\\
		\hline
	\end{tabular}
	\label{tab:EF}
\end{table}

\subsection{Task II: Identifying Species}
\label{sec:task2}

For this task, the corresponding output layer produces the probabilities of the input image being one of the 48 possible species. As is traditional in the field of computer vision, we report top-1 accuracy (is the answer correct?) and top-5 accuracy (is the correct answer in the top-5 guesses by the network?). The latter is  helpful in cases where multiple things appear in a picture, even if the ground-truth label in the dataset is only one of them. The top-5 score is also of particular interest in this work because AI can be used to help humans label data faster (as opposed to fully automating the task). In that context, a human can be shown an image and the AI's top-5 guesses. As we will report below, our best techniques identify the correct animal in the top-5 list 99.1\% of the time. Providing such a list thus could save humans the effort of finding the correct species name in a list of 48 species over 99\% of the time, although human user studies will be required to test that hypothesis.

\begin{figure*}[h!]
	\setlength{\unitlength}{0.14in}
	\centering 
	\includegraphics[width=0.9\textwidth]{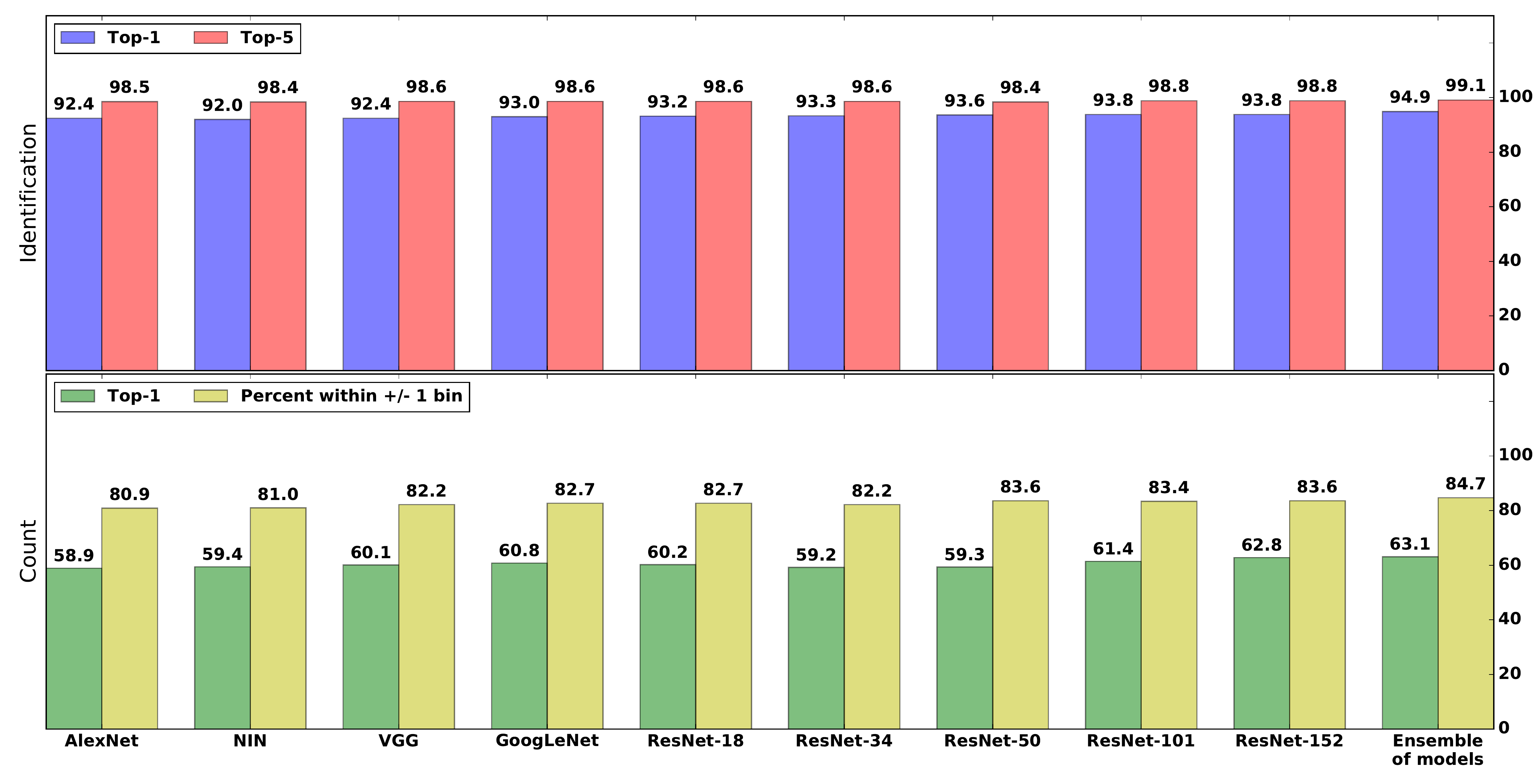}
	\caption{Top: top-1 and top-5 accuracy of different models on the task of identifying the species of animal present in the image. Although the accuracy of all the models are similar, the ensemble of models is the best with 94.9\% top-1 and 99.1\% top-5 accuracy. Bottom: top-1 accuracy and the percent of predictions within +/-1 bin for counting animals in the images. Again, the ensemble of models is the best with 63.1\% top-1 and 84.7\% of the prediction within +/-1 bin.}
	\label{fig:AccuracyResults} 
\end{figure*}

Measured on the expert-labeled test set, the model ensemble has 94.9\% top-1 and 99.1\% top-5 accuracy, while the best single model (ResNet-152) obtains 93.8\% top-1 and 98.8\% top-5 accuracy (Fig. \ref{fig:AccuracyResults} top). The results on the volunteer-labeled test set along with several examples (like Fig. \ref{fig:example}) are reported in SI Sec.~\nameref{sec:silvertestset}.

\subsection{Task III:  Counting Animals}
\label{sec:task3}
There are many different approaches for counting objects in images by deep learning \cite{Chattopadhyay_2017_CVPR,onoro2016towards,zhang2015cross} but nearly all of them require labels for bounding boxes around different objects in the image. Because this kind of information is not readily available in the SS dataset, we treat animal counting as a classification problem and leave more advanced methods for future work. In other words, instead of actually counting animals in the image, we assign the image to one of the 12 possible bins, each represents 1, 2, 3, 4, 5, 6, 7, 8, 9, 10, 11-50, or +51 individuals respectively. For this task, in addition to reporting top-1 we also report the percent of images that correctly classified within +/- 1 bins \cite{swanson2015snapshot}. 

For this task, we get 63.1\% top-1 accuracy and 84.7\% of prediction are within +/- 1 bin by the ensemble of models on the expert-labeled test set while the same metrics for the best single model (ResNet-152) are 62.8\% and 83.6\% respectively (Fig. \ref{fig:AccuracyResults} bottom). The results on the volunteer-labeled test set along with several examples are reported in SI Sec.~\nameref{sec:silvertestset}.

\subsection{Task IV: Additional Attributes}
\label{sec:task4}

The SS dataset contains labels for 6 additional attributes: standing, resting, moving, eating, interacting, and whether young are present (Fig. \ref{fig:example}). Because these attributes are not mutually exclusive (especially for images containing multiple individuals), this task is a multi-label classification \cite{tsoumakas2006multi,sorower2010literature} problem. A traditional approach for multi-label classification is to transform the task into a set of binary classification tasks \cite{tsoumakas2006multi,read2011classifier}. We do so by having, for each additional attribute, one two-neuron output layer that predicts the probability of that behavior existing (or not) in the image.

The expert-labeled test set does not contain labels for these additional attributes, so we use the majority vote among the volunteer labels as the ground truth label for each attribute. 
We count an output correct if the prediction of the model for that attribute is higher than 50\% and matches the ground-truth label. 
We report traditional multi-label classification metrics, specifically, multi-label accuracy, precision, and recall \cite{sorower2010literature}. Pooled across all attributes, the ensemble of models produces 76.2\% accuracy, 86.1\% precision, and 81.1\% recall. The same metrics for the best single model (ResNet-152) are 75.6\%, 84.5\%, and 80.9\% respectively. The full results for predicting additional attributes along with several examples are reported in SI Sec.~\nameref{sec:silvertestset}.

\section*{Saving Human Labor via Confidence Thresholding}
\label{sec:confidence}
One main benefit of automating information extraction is eliminating the need for humans to have to label images. Here we estimate the total amount of human labor that can be saved if our system is designed to match the accuracy of human volunteers.

We create a two-stage pipeline by having the VGG model from the empty vs. animal experiment classify whether the image contains an animal and, if it does, having the ensemble of models from the second stage label it. We can ensure the entire pipeline is as accurate as human volunteers by having the network classify images only if it is sufficiently confident in its prediction.

Harnessing this confidence thresholding mechanism, we can design a system that matches the volunteer human classification accuracy of 96.6\%.  For \nameref{sec:task1}, we do not have expert-provided labels and thus do not know the accuracy of the human volunteers, so we assumed it to be the same 96.6\% accuracy as on the animal identification task (Task 2). Because the VGG model's accuracy is higher than the volunteers we can automatically process 75\% of the data (because 75\% of the images are empty) at human-level accuracy. For \nameref{sec:task2}, thresholding at 43\% confidence enables us to automatically process 97.2\% of the remaining 25\% of the data at human-level accuracy. Therefore, our fully automated system operates at 96.6\% accuracy on $75\%\times100\%+97.2\%\times25\%=99.3\%$ of the data. Applying the same procedure to \nameref{sec:task3}, human volunteers are 90.0\% accurate and to match them we can threshold at 79\%. As a result, we can automatically count 44.55\% of the non-empty images and therefore $75\%*100\%+44.5\%*25\%=86.1\%$ of the data. For more details and plots please refer to SI Sec.~\nameref{sec:thresholding}. We cannot perform this exercise for \nameref{sec:task4} because SS lacks expert-provided labels for this task, meaning human-volunteer accuracy on it is unknown.

Note that to manually label $\sim$5.5 million images, nearly 30,000 SS volunteers have donated $\sim$14.6 years of 40-hour-a-week effort \cite{swanson2015snapshot}. 
Based on these statistics, \textbf{our current automatic identification system saves an estimated 8.4 years of 40-hour-per-week human labeling effort (over 17,000 hours) for 99.3\% of the 3.2 million images in our dataset}. Such effort could be reallocated to harder images  or harder problems or might enable camera-trap projects that are not able to recruit as many volunteers as the famous SS project with its charismatic megafauna.

\section*{Discussion and Future Work}

There are many directions for future work, but here we mention two particularly promising ones. 

\begin{enumerate}
\item Studying the actual time savings and effects on accuracy of a system hybridizing deep neural networks and teams of human volunteer labelers. Time savings should come from three sources: automatically filtering empty images, accepting automatically extracted information from images for which the network is highly confident in, and by providing human labelers with a sorted list of suggestions from the model so they can quickly select the correct species, counts, and descriptions. However, the actual gains seen in practice need to be quantified. Additionally, the effect of such a hybrid system on human accuracy needs to be studied. Accuracy could be hurt if humans are more likely to accept incorrect suggestions from deep neural networks, but could also be improved if the model suggests information that humans may not have thought to consider. 

\item Harnessing transfer learning to automate animal identification for camera-trap projects that do not have access to large labeled datasets. The challenge in such cases is how to train a model without access to many labeled images. Transfer learning can help, wherein a deep neural network is trained on a large, labeled dataset initially and then the knowledge learned is repurposed to classify a different dataset with fewer labeled images \cite{yosinski2014transferable}. We found that transfer learning between ImageNet and SS was not helpful (SI Sec.~\nameref{app:transfer}), but ImageNet contains many human-made categories
and the features learned to classify human-made objects (e.g. computer keyboards or Christmas ornaments) may not help when classifying animals.
Previous transfer learning research has shown that it works better the more similar the transfer-from and transfer-to tasks are \cite{yosinski2014transferable}. 
Transferring from one animal dataset to another one may prove more fruitful. Experiments need to be conducted to verify the extent to which transfer learning from the SS dataset or others can help automate knowledge extraction from other camera-trap projects with fewer labeled images.
\end{enumerate}
\section*{Conclusions}
In this paper, we tested the ability of state-of-the-art computer vision methods called deep neural networks to automatically extract information from images in the SS dataset, the largest existing labeled dataset of wild animals. We first showed that deep neural networks can perform well on the SS dataset, although performance is worse for rare classes. 

Perhaps most importantly, our results show that employing deep learning technology can save a tremendous amount of time for biology researchers and the human volunteers that help them by labeling images. In particular, for animal identification, our system can save 99.3\% of the manual labor (over 17,000 hours) while performing at the same 96.6\% accuracy level of human volunteers. This substantial amount of human labor can be redirected to other important scientific purposes and also makes knowledge extraction feasible for camera-trap projects that cannot recruit large armies of human volunteers.
Automating data extraction can thus dramatically reduce the cost to extract valuable information from wild habitats, likely revolutionizing studies of animal behavior, ecosystem dynamics, and wildlife conservation. 

\showmatmethods 

\acknow{Jeff Clune was supported by an NSF CAREER award (CAREER: 1453549). 
All experiments were conducted on the Mount Moran IBM System X cluster computer at the University of Wyoming Advanced Research Computing Center (ARCC). The authors thank the ARCC staff for their support, and the members of the Evolving AI Lab at the University of Wyoming for valuable feedback on this draft, especially Joost Huizinga, Tyler Jaszkowiak, Roby Velez, and Nick Cheney. We also thank the Snapshot Serengeti volunteers \url{https://www.snapshotserengeti.org/\#/authors}.}

\showacknow 

\bibliography{master}
\clearpage 

\twocolumn[{
	\centering
	\titlefont\textbf{Supplementary Information}
	\normalsize
	\vspace*{0.8cm}
}]
\date{}
\author{}
\renewcommand\thefigure{S.\arabic{figure}}
\renewcommand\thesection{S.\arabic{section}}
\renewcommand\thetable{S.\arabic{table}}
\setcounter{section}{0}
\setcounter{figure}{0}
\setcounter{table}{0}


\section{Pre-processing and Training}
\label{app:archs}

In this section, we document the technical details for the pre-processing step and for selecting the hyperparameters across all experiments in the paper.

\subsection*{Pre-processing} 
\begin{figure}[h]
	\setlength{\unitlength}{0.34in}
	\centering 
	\includegraphics[width=0.5\textwidth]{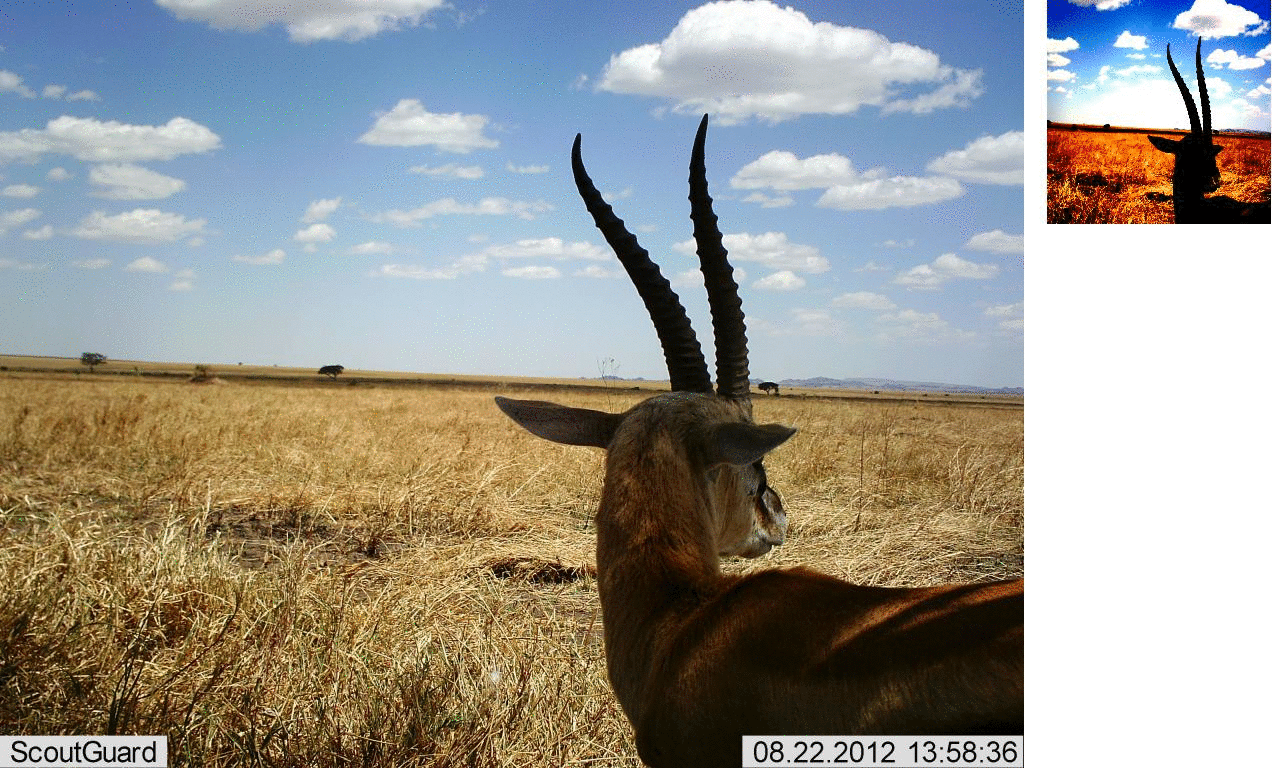}
	\caption{An example of a camera-trap image in the SS dataset (left) and its down-sampled, normalized equivalent (upper right), which is what is actually input to the neural network.}
	\label{fig:preprocess} 
\end{figure}

The original images in the dataset are 2,048$\times$1,536 pixels, which is too large for current state-of-the-art deep neural networks owing to the increased computational costs of training and running DNNs on high-resolution images. We followed standard practices in scaling down the images to 256$\times$256 pixels. Although this may distort the images slightly, since we do not preserve the aspect ratios of the images, it is a de facto standard in the deep learning community \cite{Goodfellow-et-al-2016-Book}. The images in the dataset are color images, where each pixel has three values: one for each of the red, green, and blue intensities. We refer to all the values for a specific color as a color channel. After scaling down the images, we computed the mean and standard deviation of pixel intensities for each color channel separately and then we normalized the images by subtracting the average and dividing by the standard deviation (Fig. \ref{fig:preprocess}). This step is known to make learning easier for neural networks \cite{lecun2012efficient,wiesler2011convergence}. 

\subsection*{Data Augmentation} We perform random cropping, horizontal flipping, brightness modification, and contrast modification to each image. Doing so, we provide an slightly different image each time, which can make the network resistant to small changes and improve the accuracy of the network \cite{krizhevsky2012imagenet}.

\subsection*{Training} We train the networks via backpropagation using Stochastic Gradient Descent (SGD) optimization with momentum and weight decay \citep{Goodfellow-et-al-2016-Book}. We used the Torch \cite{collobert2002torch} and Tensorflow \cite{tensorflow2015-whitepaper} frameworks for our experiments. The SGD optimization algorithm requires several hyperparameters. The settings for those in our experiments are in Table \ref{app:tab:hyp}. We train each model for 55 epochs with the learning-rate policy and the weight-decay policy that are shown in Table \ref{app:tab:lr}. We checkpoint the model after each epoch and at the end, we report the results of the most accurate model on the expert-labeled test set.

\begin{table}[h!]
	\caption{The static neural network training hyperparameters for all experiments.}
	\centering
	\begin{tabular}{cc}
		\textbf{Hyperparameter}&\textbf{Value}\\
		\hline
		Batch Size&128\\
		Momentum&0.9\\
		Crop Size&224$\times$224\\
		Number of Epochs&55\\
		Epoch Size&5900\\
		\hline
	\end{tabular}
	\label{app:tab:hyp}
\end{table}

\begin{table}[h]
	\caption{The dynamic neural network training hyperparameters for all experiments.}
	\centering
	\begin{tabular}{ccc}
		\textbf{Epoch Number}&\textbf{Learning Rate}&\textbf{Weight Decay}\\
		\hline
		1-18&0.01&0.0005\\
		19-29&0.005&0.0005\\
		30-43&0.001&0\\
		44-52&0.0005&0\\
		53&0.0001&0\\
		\hline
	\end{tabular}
	\label{app:tab:lr}
\end{table}

\section{One-stage Identification}
\label{sec:onestep}
In the main text, we employ a two-step pipeline for automatically processing the camera-trap images. The first step tries to filter out empty images and the second step provides information about the remaining images. One possibility is merging these two steps into just one step. We can consider the empty images as one of the identification classes and then train models to classify input images either as one of the species or the empty class. Although this approach results in a smaller total model size than having separate models for the first and second steps, there are three drawbacks to this approach. (a) Because around 75\% of the images are empty images, this approach imposes a great deal of imbalance between the empty and other classes, which makes the problem harder for machine learning algorithms. (b) A one-step pipeline does not enable us to reuse an empty vs. animal module for other similar datasets. (c) We find out that one-step pipeline produces slightly worse results. In our experiment, to avoid the imbalance issue, we randomly select 220,000 empty images for the empty class, which is equal to the number of images for the most frequent class (wildebeest). Then we train four different architectures and measure their total accuracy, empty vs. animal accuracy, and species identification accuracy. The results are shown in Table \ref{app:tab:onestep}.

\begin{table}[h]
	\caption{The results of the one-stage identification experiment. Although one-stage models do produce good results, their results are slightly worse than their corresponding two-stage comparator. For example, on \nameref{sec:task1}, the one-step ResNet-50 model has 94.9\% accuracy vs. 96.3\% for the two-stage pipeline. For \nameref{sec:task2} the one-step ResNet-50 is 90.6\% accurate with a one-step model vs. 93.6\% for the two-stage pipeline.}
	\centering
	\begin{tabular}{p{1.5cm}p{2cm}p{2.2cm}p{1.5cm}}
		\textbf{Architecture}&\textbf{Total Accuracy}&\textbf{Empty vs. Animal Accuracy}&\textbf{Identification Accuracy}\\
		\hline
		AlexNet&88.9\%&93.7\%&87.9\%\\
		ResNet-18&90.5\%&95.4\%&89.5\%\\
		ResNet-34&90.8\%&94.7\%&90.0\%\\
		ResNet-50&91.3\%&94.9\%&90.6\%\\
		\hline
	\end{tabular}
	\label{app:tab:onestep}
\end{table}

\section{Results on the Volunteer-Labeled Test Set}
\label{sec:silvertestset}
\begin{figure*}[h!]
	\setlength{\unitlength}{0.14in}
	\centering 
	\includegraphics[width=0.9\textwidth]{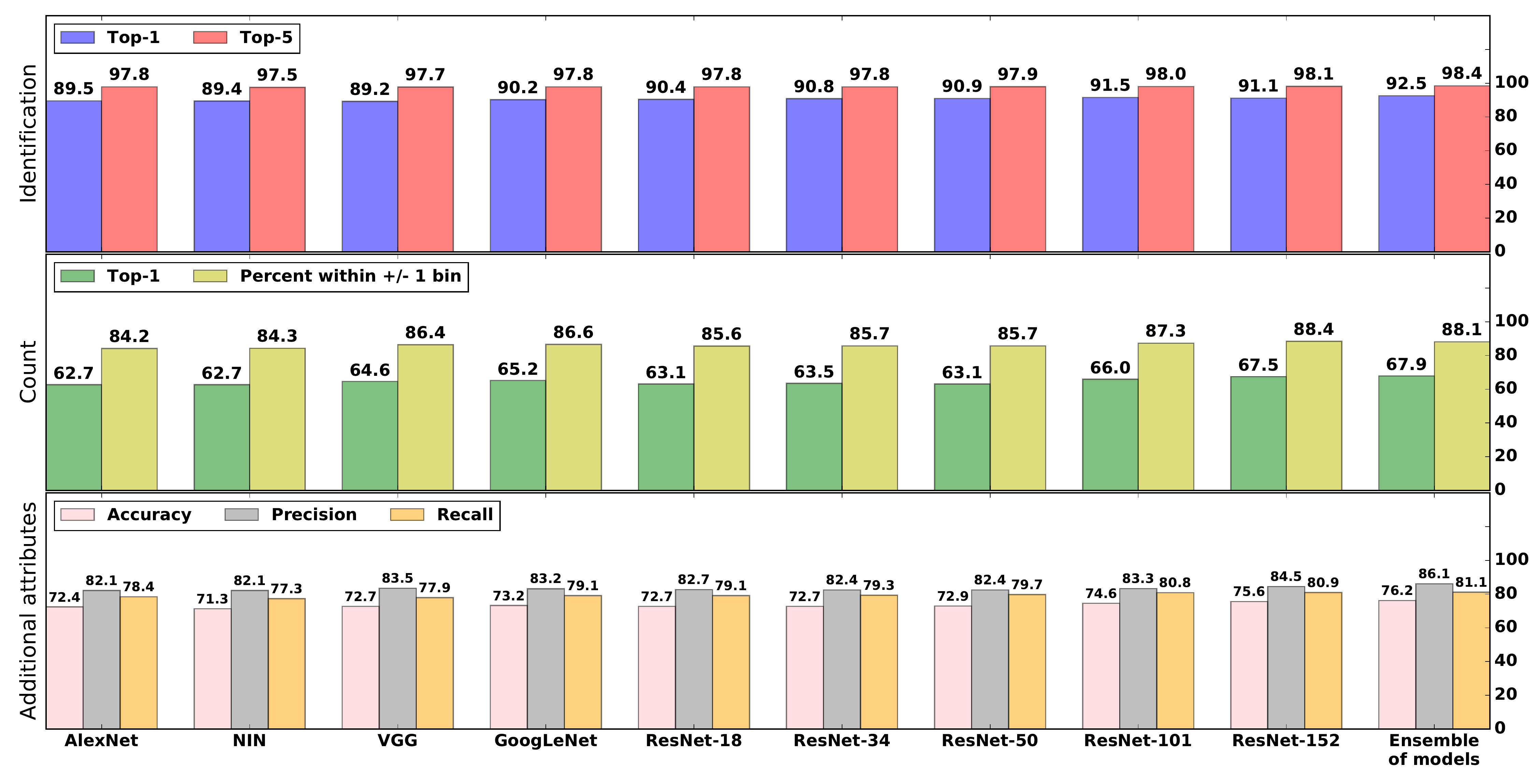}
	\caption{The results of \nameref{sec:task2}, \nameref{sec:task3}, and \nameref{sec:task4} on the volunteer-labeled test set. The top plot shows top-1 and top-5 accuracy of different models for the task of identifying animal species. The ensemble of models is the best with 92.5\% top-1 accuracy and 98.4\% top-5 accuracy. The middle plot shows top-1 accuracy and the percent of predictions within +/-1 bin for counting animals in the images. The ensemble of models has the best top-1 accuracy with 67.9\% and ResNet-152 has the closest predictions with 88.4\% of the prediction within +/-1 bin. The bottom plot shows accuracy for the task of describing additional attributes (behaviors and the presence of young). The ensemble of models is the best with 76.2\% accuracy, 86.1\% precision, and 81.1\% recall.}
	\label{fig:SilverResults} 
\end{figure*}

As mentioned in the main text, the volunteer-labeled test set has 17,400 capture events labeled by human volunteers. It has labels for species, counts, descriptions of animal behaviors, and whether young are present. In the main paper we compared our model predictions to expert-provided labels; in this section we compare instead to the volunteer-provided labels. Fig. \ref{fig:SilverResults} shows the results. For \nameref{sec:task2}, all the models have top-1 accuracy above 89.2\% and top-5 accuracy above 97.5\%. For \nameref{sec:task3}, all models have top-1 accuracy more than 62.7\% and all of them can count within one bin for over 84.2\% of the test examples. 

\begin{figure*}
	\centering
	\subfloat[]{{\includegraphics[width=0.30\textwidth]{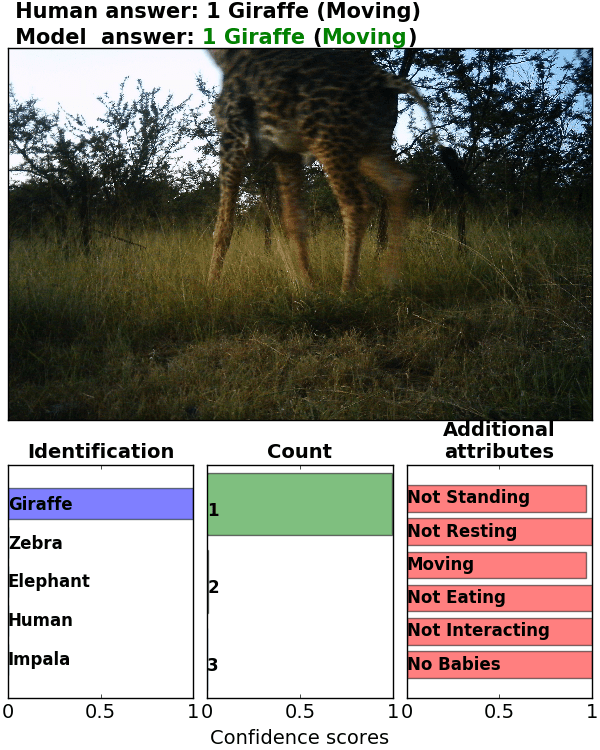} }}%
	\subfloat[]{{\includegraphics[width=0.30\textwidth]{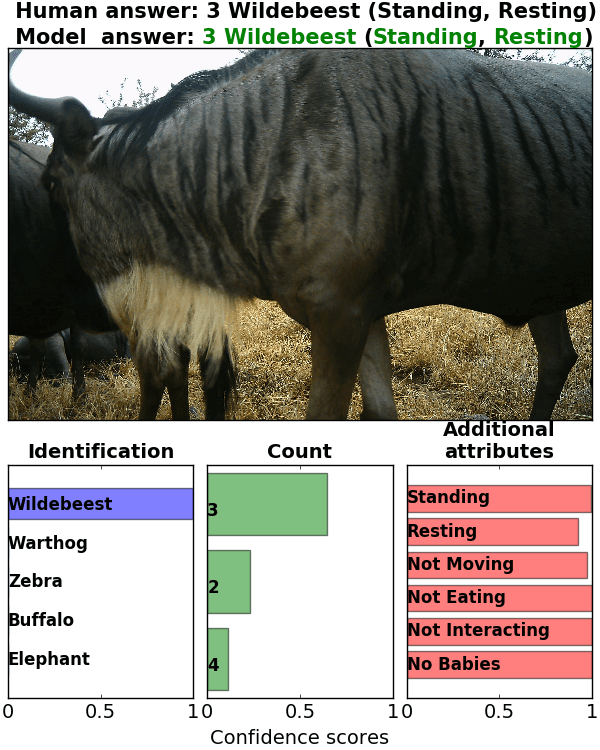} }}%
	\subfloat[]{{\includegraphics[width=0.30\textwidth]{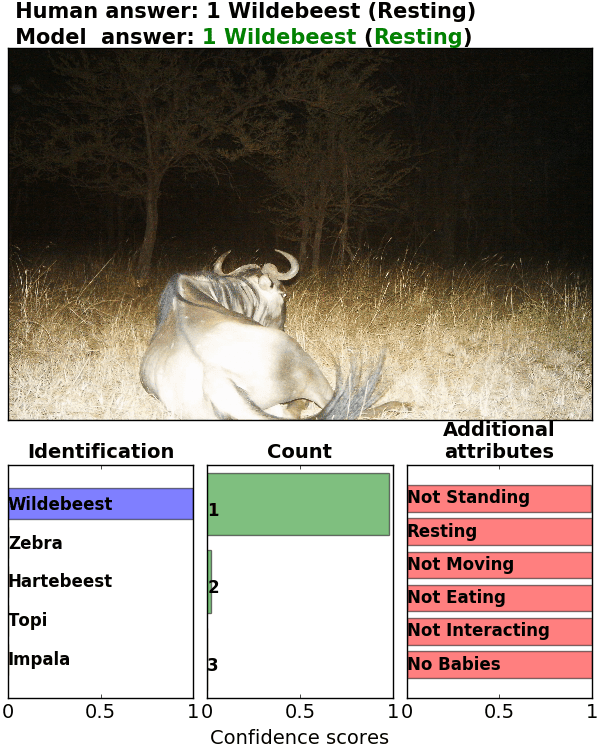} }}%
	\hskip 5pt 	
		\subfloat[]{{\includegraphics[width=0.30\textwidth]{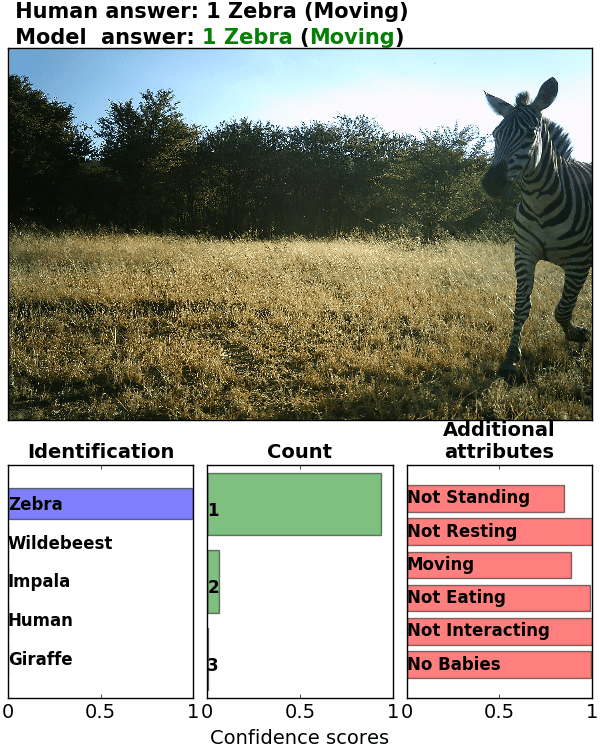} }}%
	\subfloat[]{{\includegraphics[width=0.30\textwidth]{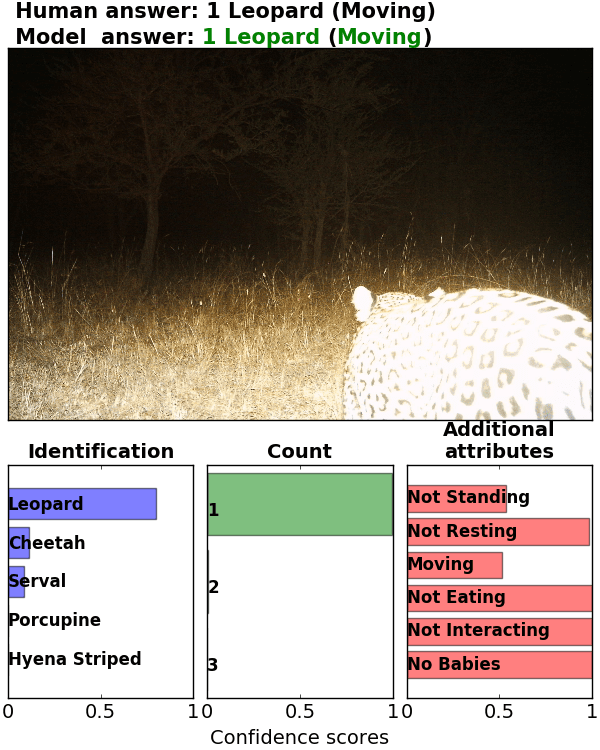} }}%
	\subfloat[]{{\includegraphics[width=0.30\textwidth]{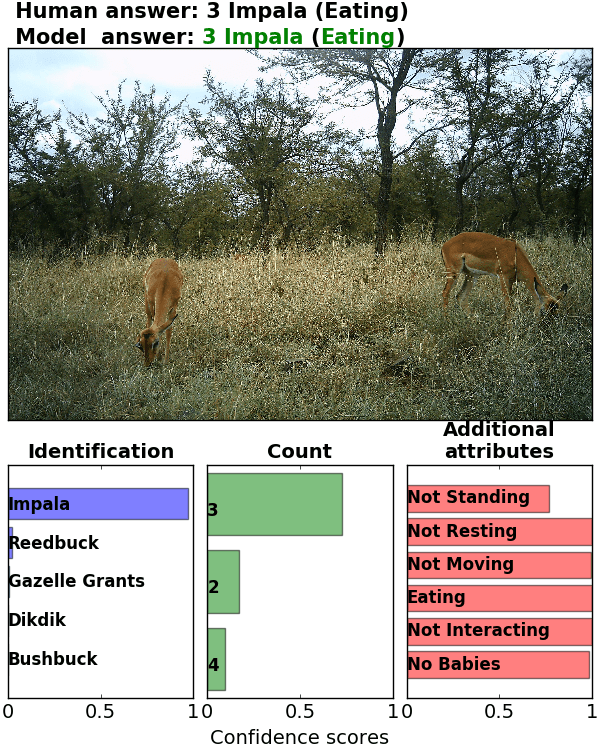} }}%
	\hskip 5pt 
		\subfloat[]{{\includegraphics[width=0.30\textwidth]{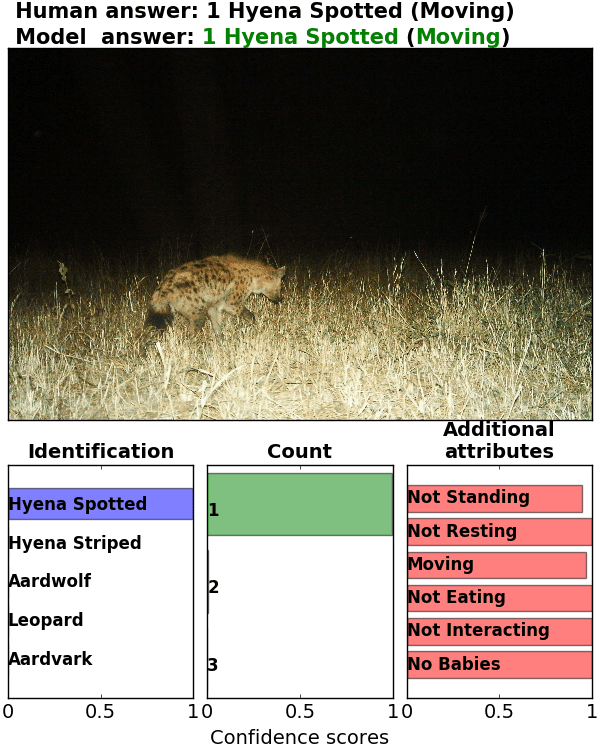} }}%
	\subfloat[]{{\includegraphics[width=0.30\textwidth]{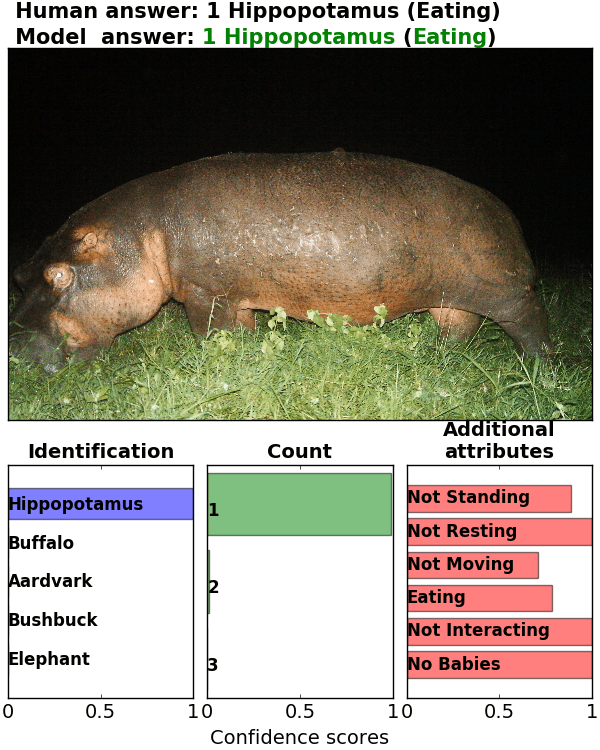} }}%
	\subfloat[]{{\includegraphics[width=0.30\textwidth]{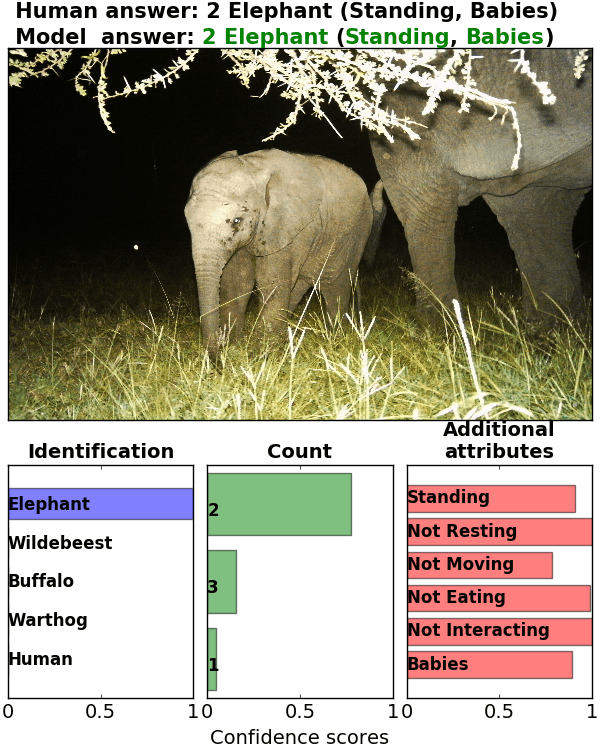} }}%
	\caption{Shown are nine images the ResNet-152 model labeled correctly. Above each image are a combination of expert-provided labels (for the species type and counts) and volunteer-provided labels (for additional attributes), as well as the model's prediction for that image. Below each image are the top guesses of the model for different tasks, with the width of the color bars indicating the model's output for each of the guesses, which can be interpreted as its confidence in that guess.	}
	\label{fig:corrects}
\end{figure*}

\begin{figure*}
	\centering
	\subfloat[]{{\includegraphics[width=0.30\textwidth]{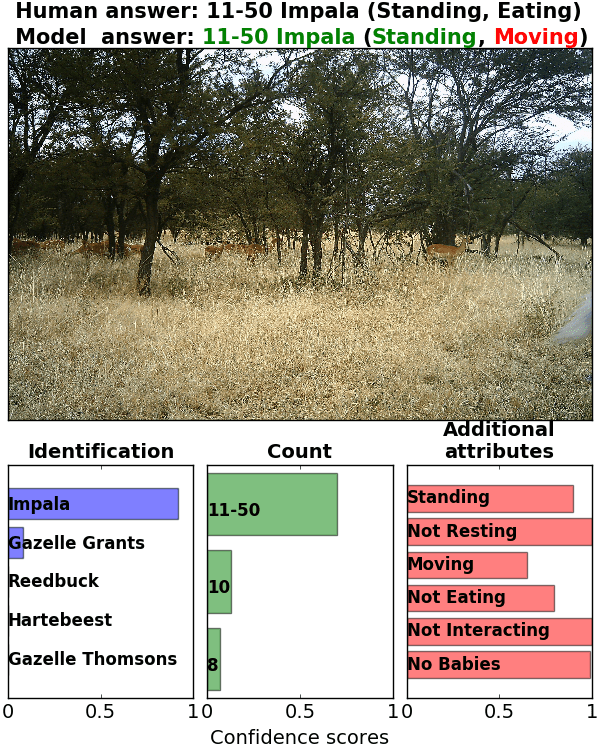} }}%
	\subfloat[]{{\includegraphics[width=0.30\textwidth]{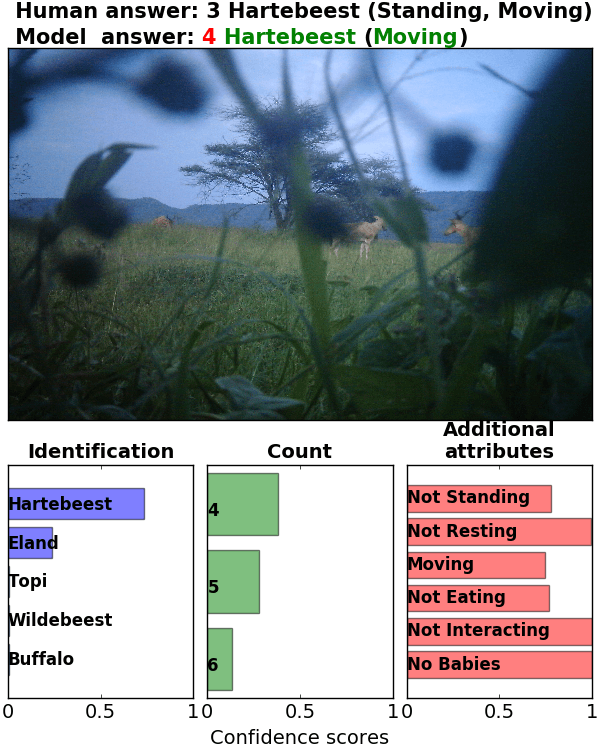} }}%
	\subfloat[]{{\includegraphics[width=0.30\textwidth]{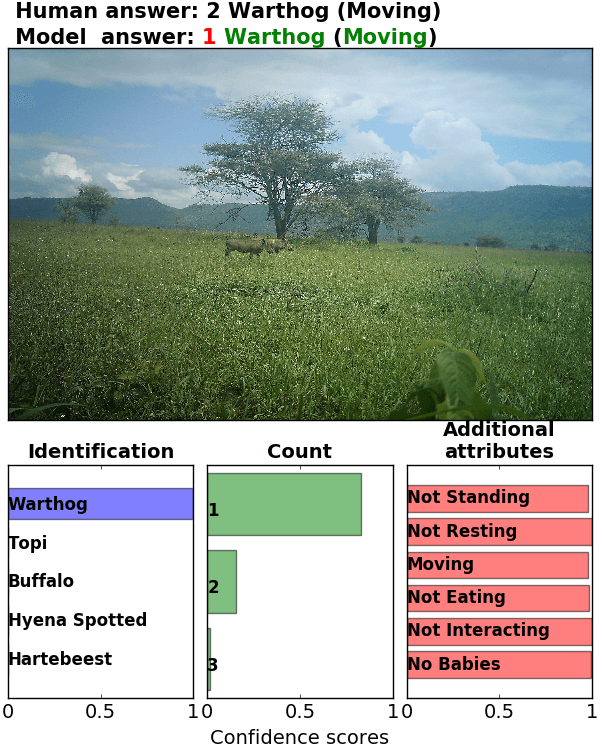} }}%
	\hskip 5pt 	
		\subfloat[]{{\includegraphics[width=0.30\textwidth]{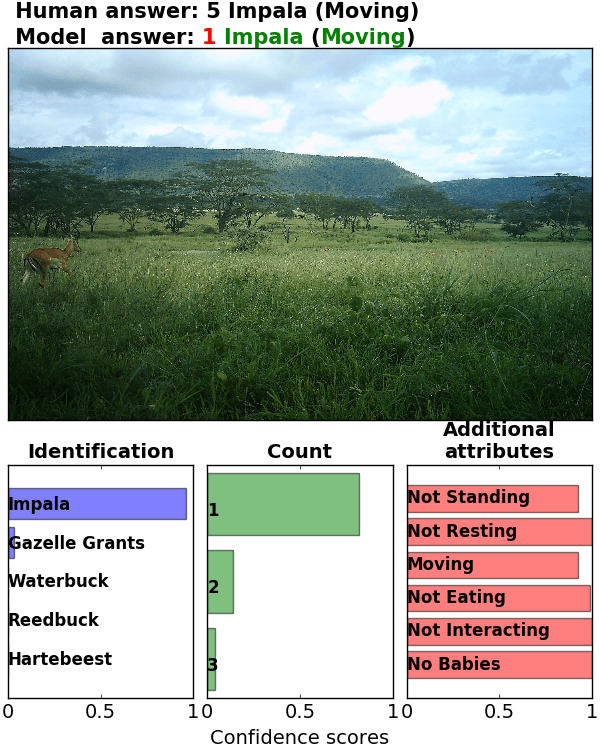} }}%
	\subfloat[]{{\includegraphics[width=0.30\textwidth]{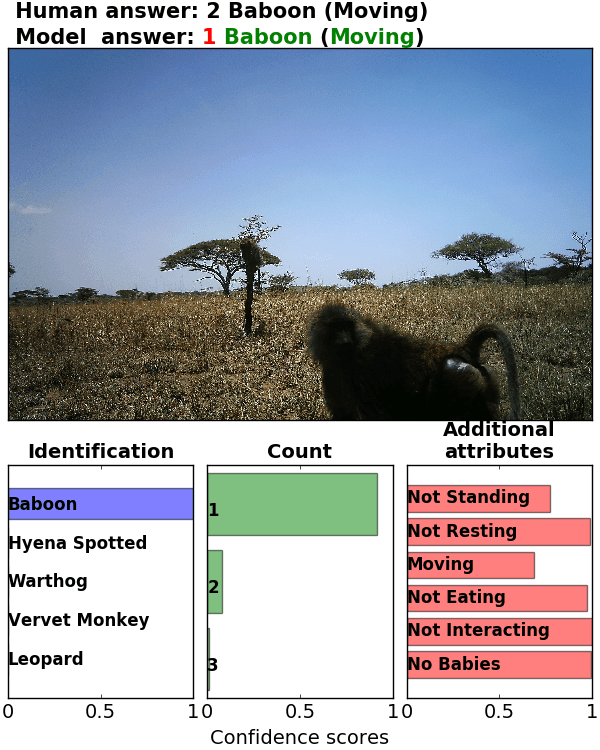} }}%
	\subfloat[]{{\includegraphics[width=0.30\textwidth]{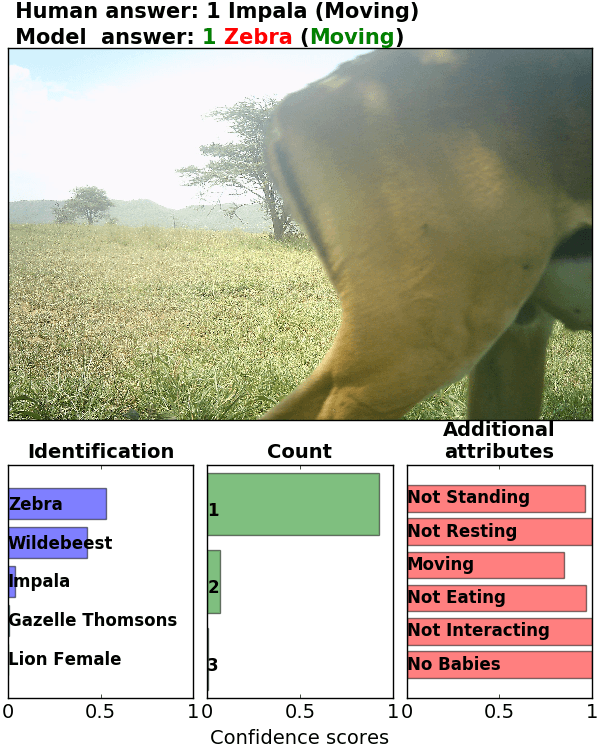} }}%
	\hskip 5pt 
		\subfloat[]{{\includegraphics[width=0.30\textwidth]{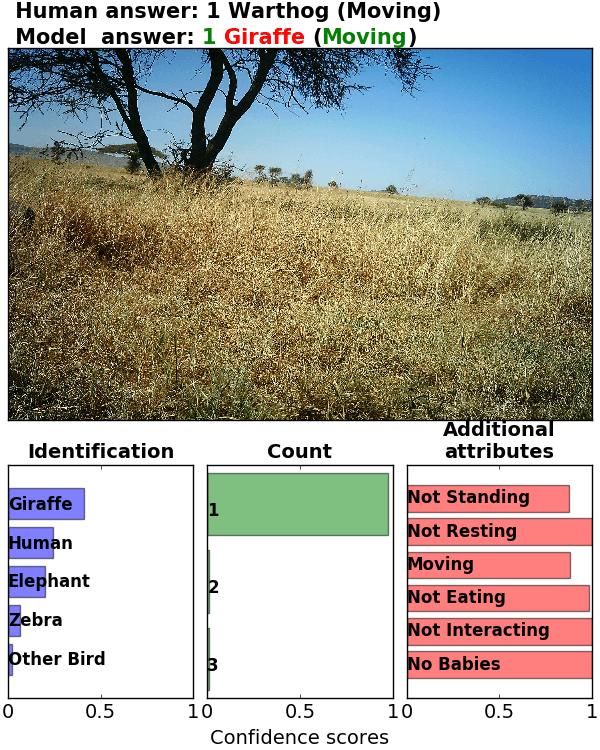} }}%
	\subfloat[]{{\includegraphics[width=0.30\textwidth]{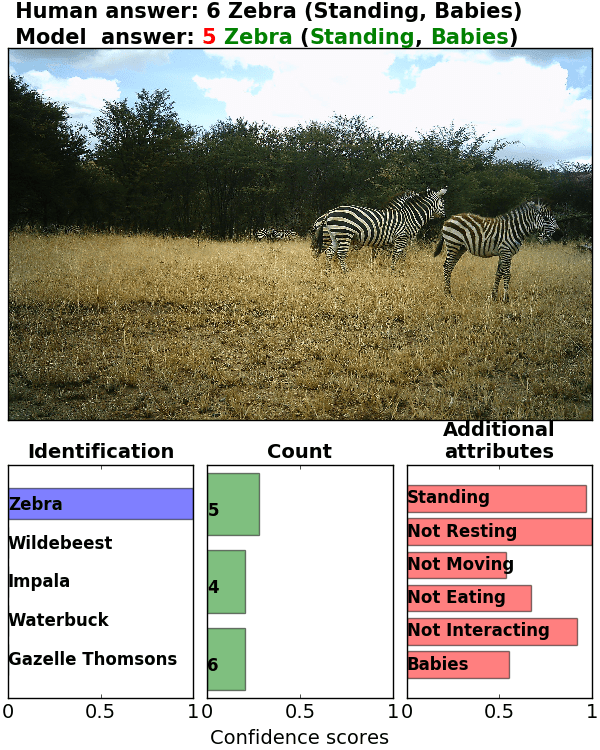} }}%
	\subfloat[]{{\includegraphics[width=0.30\textwidth]{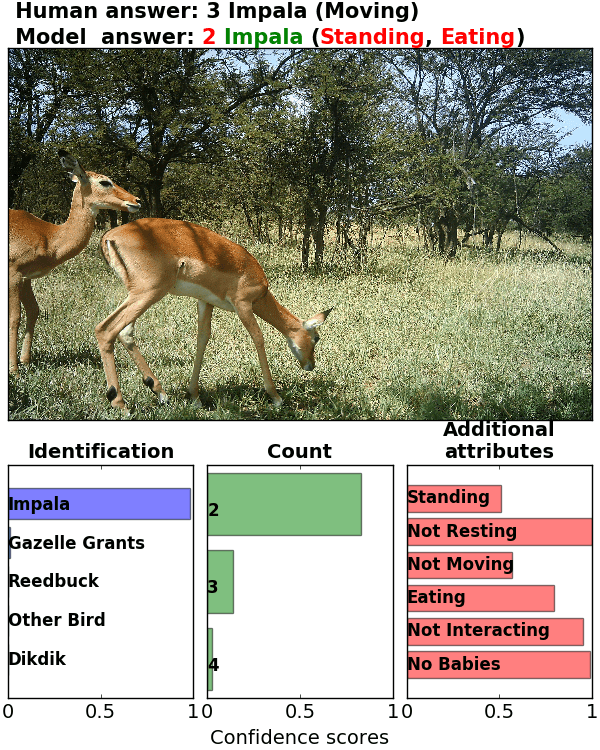} }}%
	\caption{Shown are nine images the ResNet-152 model labeled incorrectly. Above each image are a combination of expert-provided labels (for the species type and counts) and volunteer-provided labels (for additional attributes), as well as the model's prediction for that image. Below each image are the top guesses of the model for different tasks, with the width of the color bars indicating the model's output for each of the guesses, which can be interpreted as its confidence in that guess. One can see why the images are difficult to get right. \textbf{(g, i)} contain examples of the noise caused by assigning the label for the capture event to all images in the event. \textbf{(a, b, d, h)} show how animals being too far from the camera makes classification difficult.}
	\label{fig:mistakes}
\end{figure*}

For \nameref{sec:task4}, the models have at least 71.3\% accuracy, 82.1\% precision, and 77.3\% recall. The ensemble of models performs the best for the description task by a small margin.
Overall, for all the tasks, the results of different architectures are similar. Moreover, our models predictions are closer to those of the experts on some tasks (e.g. animal identification), and closer to human-volunteers on others (e.g. counting), for reasons that are not clear.
We provide examples of correct predictions (Fig. \ref{fig:corrects}) and partially or fully incorrect network predictions (Fig. \ref{fig:mistakes}).

\section{Comparing to Gomez et al. 2016}
\label{sec:animalIdentificationCOLO} 
In the closest work to ours, Gomez et al. \cite{gomez2016towards} employed \emph{transfer learning} \cite{yosinski2014transferable,torrey2009transfer}, which is a way to learn a new task by utilizing knowledge from an already learned, related task. In particular, they used models pre-trained on the ImageNet dataset, which contains 1.3 million images from 1,000 classes of man-made and natural images \cite{deng2009imagenet} to extract features and then, on top of these high-level features, trained a linear classifier to classify animal species. They tested six different architectures: AlexNet \cite{krizhevsky2012imagenet}, VGG \cite{simonyan2014very}, GoogLeNet \cite{szegedy2015going}, ResNet-50 \cite{he2015deep}, ResNet-101 \cite{he2015deep}, and ResNet-152 \cite{he2015deep}. To improve the results for two of these architectures, they also further trained the entire AlexNet and GoogLeNet models on the SS dataset (a technique called fine-tuning \citep{yosinski2014transferable,torrey2009transfer,Goodfellow-et-al-2016-Book}).

To avoid dealing with an unbalanced dataset, Gomez et al. \cite{gomez2016towards} removed all species classes that had a small number of images and classified only 26 out of the total 48 SS classes. Because we want to compare our results to theirs and since the exact dataset used in \cite{gomez2016towards} is not publicly available, we did our best to reproduce it by including all images from those 26 classes. We call this dataset SS-26. We split 93\% of the images in SS-26 into the training set and place the remaining 7\% into the test set (the training vs. test split was not reported in Gomez et al. \cite{gomez2016towards}). 

\begin{figure}[h!]
	\setlength{\unitlength}{0.14in}
	\centering 
	\includegraphics[width=0.5\textwidth]{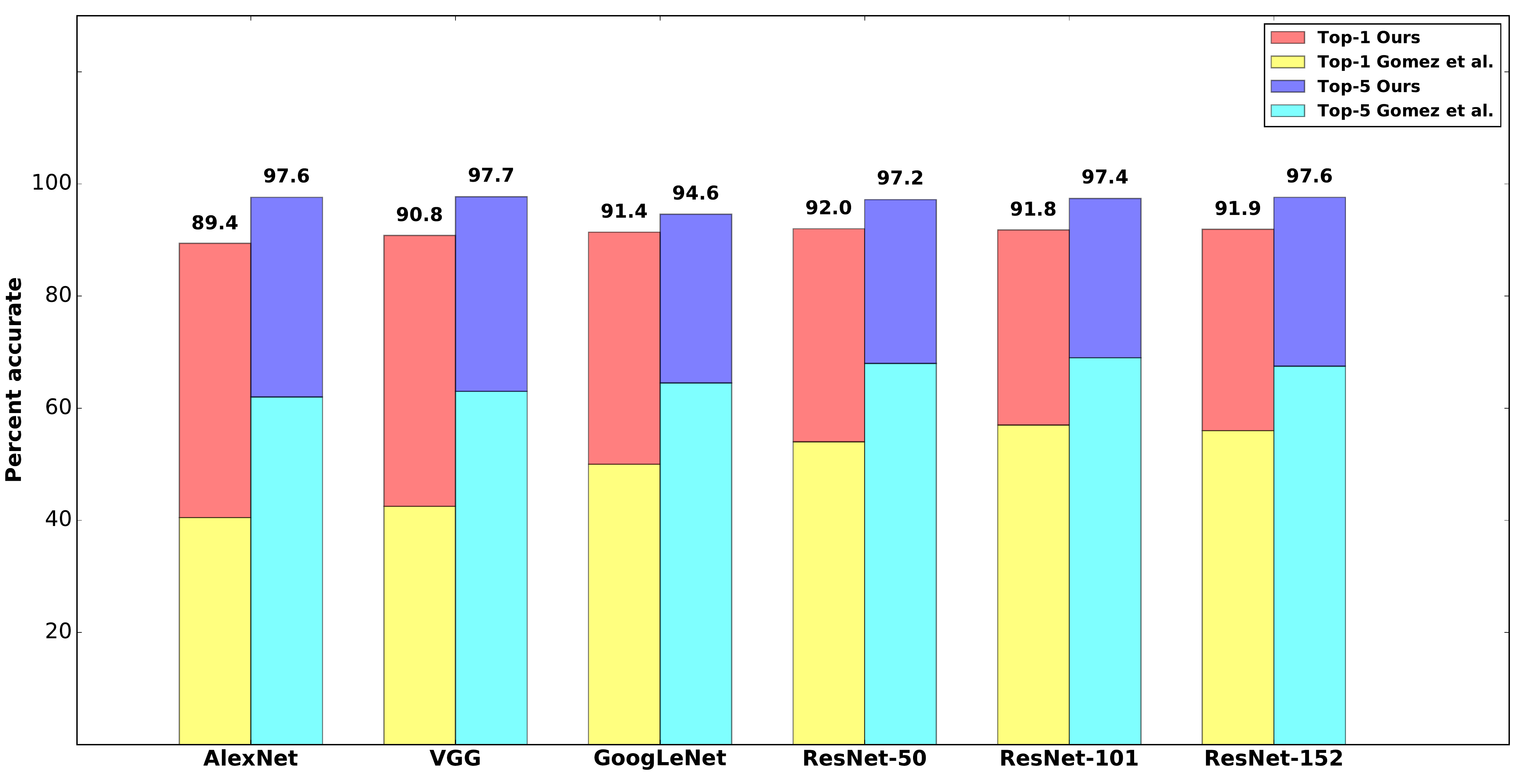}
	\caption{For the experiment classifying the 26 most common species, shown is the top-1 and top-5 accuracy from Gomez et al. \cite{gomez2016towards} and for the different architectures we tested. Our models yield significantly better results. On average, top-1 and top-5 accuracies are improved over 30\%. The ResNet-50 model achieved the best top-1 result with 92\% accuracy. Because Gomez et al. \cite{gomez2016towards} did not report exact accuracy numbers, the numbers used to generate this plot are estimated from their plot.}
	\label{fig:AccuracyResultsCOLO} 
\end{figure}

Because we found transfer learning from ImageNet not to help on identifying animals in the SS dataset (SI Sec. \nameref{app:transfer}), we train our networks from scratch on the SS-26 dataset. We train the same set of network architectures (with just one output layer for the identification task) as in Gomez et al. \cite{gomez2016towards} on the SS-26 dataset. For all networks, we obtained substantially higher accuracy scores than those reported in \cite{gomez2016towards} (Fig. \ref{fig:AccuracyResultsCOLO}): our best network obtains a top-1 accuracy of 92.0\% compared to around 57\% by Gomez et al. (estimating from their plot, as the exact accuracy was not reported). It is not clear why the performance of Gomez et al. \cite{gomez2016towards} is lower.

In another experiment, Gomez et al. \cite{gomez2016towards} obtained a higher accuracy of 88.9\%, but on another heavily simplified version of the SS dataset. This modified dataset contains only $\sim$33,000 images and the images were manually cropped and specifically chosen to have animals in the foreground \cite{gomez2016towards}. We instead seek deep learning solutions that perform well on the full SS dataset and without manual intervention.

\section{Transfer Learning}
\label{app:transfer}

Transfer learning \cite{pan2010survey,yosinski2014transferable} takes advantage of the knowledge gained from learning on one task and applies it to a different, related task. Our implementation of transfer learning follows from other works in the image recognition field \cite{oquab2014learning,donahue2014decaf,sharif2014cnn}. We first pre-train the AlexNet and ResNet-152 architectures on the ImageNet dataset \cite{deng2009imagenet}. These pre-trained models then become the starting point (i.e. initial weights) for training the models on the SS dataset. The static and dynamic hyperparameters for these runs are the same as in the original experiment (\nameref{app:archs}).

At the end of transfer learning, for \nameref{sec:task2}, the AlexNet model has 92.4\% top-1 accuracy and 98.8\% top-5 accuracy, while the ResNet-152 model has 93.0\% top-1 accuracy and 98.7\% top-5 accuracy. For \nameref{sec:task3}, Alexnet and ResNet-152  are 59.1\% and 62.4\% top-1 accurate and 80.7\% and 82.6\% of their predictions are only 1 bin off, respectively.

Comparing the obtained results to those in Fig.~\ref{fig:AccuracyResults} indicates that transfer learning from ImageNet does not help to increase accuracy. Although transfer learning was ineffective in these experiments, perhaps it would perform well with different hyperparameters (e.g. different learning rates).
\section{Prediction Averaging}
\label{app:merge}
For each image, a model outputs a probability distribution over all classes. For each class, we average the probabilities from the $m$ models, and then either take the top class or top $n$ classes in terms of highest average(s) as the prediction(s). Table \ref{app:tab:exm} shows an example. 

\begin{table*}[h]
	\caption{An example of classification averaging. The numbers are the probability the network estimates the input was of that class, which can also be interpreted as the network's confidence in its prediction. For all classes (e.g. species in this example), we average these confidence scores across all the models. The final aggregate prediction is the class with the highest average probability (or the top $n$ if calculating top-$n$ accuracy). Due to space constraints, we show the top 7 species (in order) in terms of average probability.}
	\centering
	\begin{tabular}{ccccc}
		\textbf{Species}&\textbf{Network 1}&\textbf{Network 2}&\textbf{Network 3}&\textbf{Average Probability}\\
		\hline
		Zebra& 0.80&0.05&0.50&(0.80+0.05+0.50)/3= 0.45\\
		Impala&0.00&0.90&0.08&(0.00+0.90+0.08)/3= 0.33\\
		Topi&0.10&0.00&0.40&(0.10+0.00+0.40)/3= 0.17\\
		Dikdik&0.07&0.04&0.00&(0.07+0.04+0.00)/3= 0.04\\
		Reedbuck&0.03&0.00&0.02&(0.03+0.00+0.02)/3= 0.02\\
		Gazelle Grants&0.00&0.01&0.00&(0.00+0.01+0.00)/3= 0.00\\
		Eland&0.00&0.00&0.00&(0.00+0.00+0.00)/3= 0.00\\
		\hline
	\end{tabular}
	\label{app:tab:exm}
\end{table*}

\section{Classifying Capture Events}
\label{sec:capture_events}
The SS dataset contain labels for \emph{capture events}, not individual images. However, our DNNs are trained to classify images. We can aggregate the predictions for individual images to predict the labels for entire capture events. One could also simply train a neural network to directly classify capture events. We try both of these approaches and report the results here.

\begin{figure}
	\setlength{\unitlength}{0.14in}
	\centering 
	\includegraphics[width=0.5\textwidth]{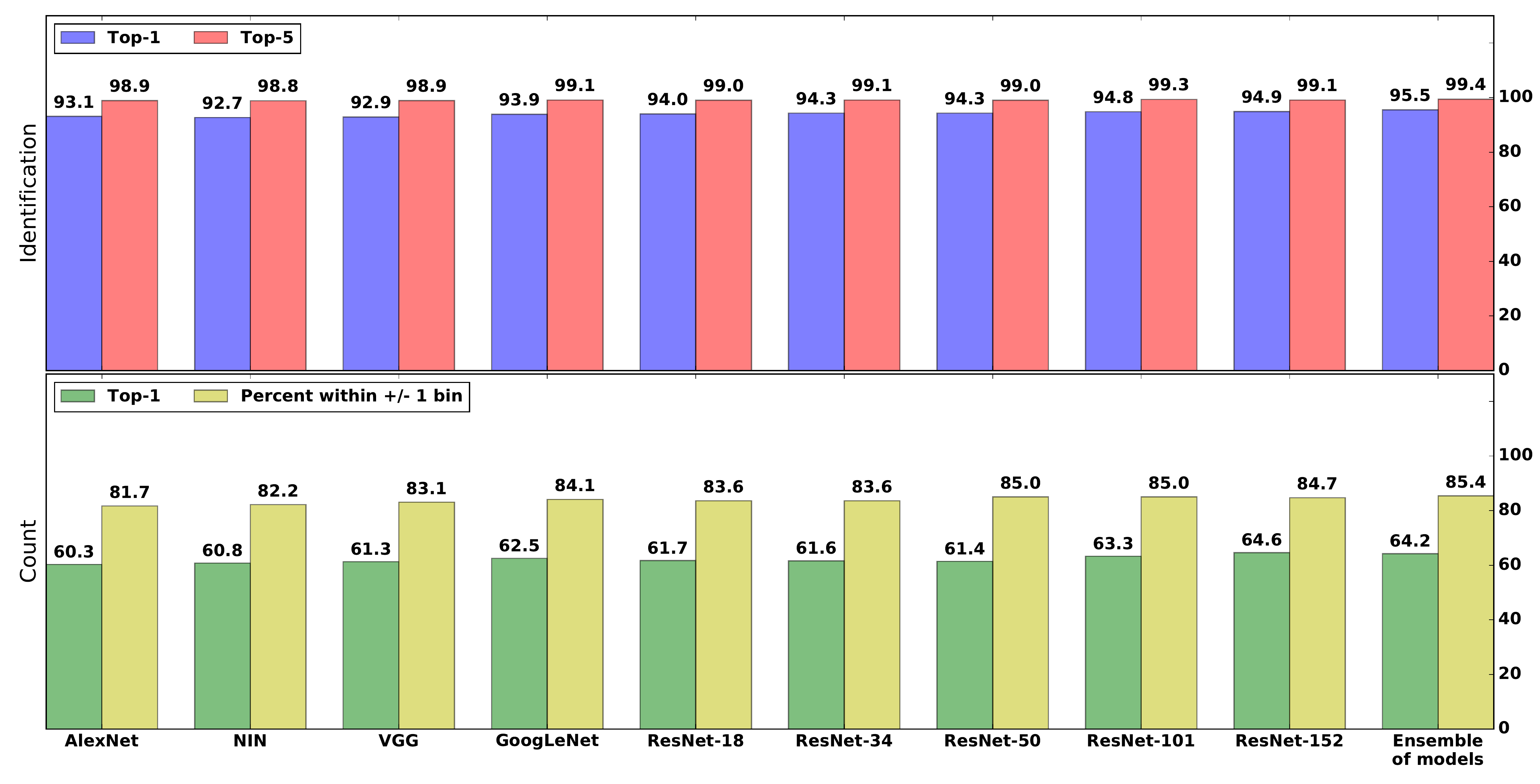}
	\caption{The top-1 and top-5 accuracy of different architectures for entire capture events (as opposed to individual images) on the expert-labeled test set. Combining the classification for all the images within a capture event improves accuracy for all the models. The best accuracy belongs to the ensemble of models with 95.5\% top-1 accuracy and 99.4\% top-5 accuracy.}
	\label{fig:AccuracyResultsCE} 
\end{figure}

To implement the former, we employ the same prediction averaging method as in Sec.~\nameref{app:merge} except that in this case the classifications come from the same model, but for different images within a capture event. The resultant accuracy scores for capture events are on average 1\% higher than those for individual images (Table \ref{tab:EFCE} and Fig. \ref{fig:AccuracyResultsCE}). This performance gain is likely because averaging over all the images in a capture event can mitigate the noise introduced by deriving the training labels of individual images from capture event labels (Fig. \ref{fig:CENoise}).

The next experiment we tried was inputting all images from a capture event at the same time and asking the model to provide one label for the entire capture event. For computational reasons, we train only one of our high-performing models (ResNet-50). Because feedforward neural networks have a fixed number of inputs, we only consider capture events that contain exactly three images and we ignore the other 55,000 capture events. We put the three images from a capture event on top of each other and form a 9-channel input image for the model. On the expert-labeled dataset, the model achieved 90.8\% top-1 accuracy and 97.4\% top-5 accuracy for identification and 58.5\% top-1 accuracy and 81.1\% predictions within +/- 1 bins for counting. Both scores are slightly below our results for any of the models trained on individual images. These results and those from the previous experiment suggest that training on individual images is quite effective and produces more accurate results.

There are other reasons to prefer classifying single images. Doing so avoids (a) the challenge of dealing with capture events with different numbers of images, (b) making the number of labeled training examples smaller (which happens when images are merged into capture events), (c) the larger neural network sizes required to process many images at once, and (d) choices regarding how best to input all images at the same time to a feedforward neural network. Overall, investigating the best way to harness the extra information in multi-image capture events, and to what extent doing so is helpful vs. classifying individual images, is a promising area of future research. 

\begin{table}
	\caption{The accuracy of models for \nameref{sec:task1} on capture events.}
	\centering
	\setlength{\unitlength}{0.10in}
	\begin{tabular}{p{2cm}p{2.5cm}}
		\textbf{Architecture}&\textbf{Top-1 accuracy for capture events}\\
		\hline
		AlexNet&96.3\%\\
		NiN&96.6\%\\
		VGG&96.8\%\\
		GoogLeNet&96.9\%\\
		ResNet-18&96.8\%\\
		ResNet-34&96.8\%\\
		ResNet-50&97.1\%\\
		ResNet-101&96.8\%\\
		ResNet-152&96.8\%\\
		\hline
	\end{tabular}
	\label{tab:EFCE}
\end{table}

\section{Confidence Thresholding}
\label{sec:thresholding}
\begin{figure}[h!]
	\setlength{\unitlength}{0.14in}
	\centering 
	\includegraphics[width=.5\textwidth]{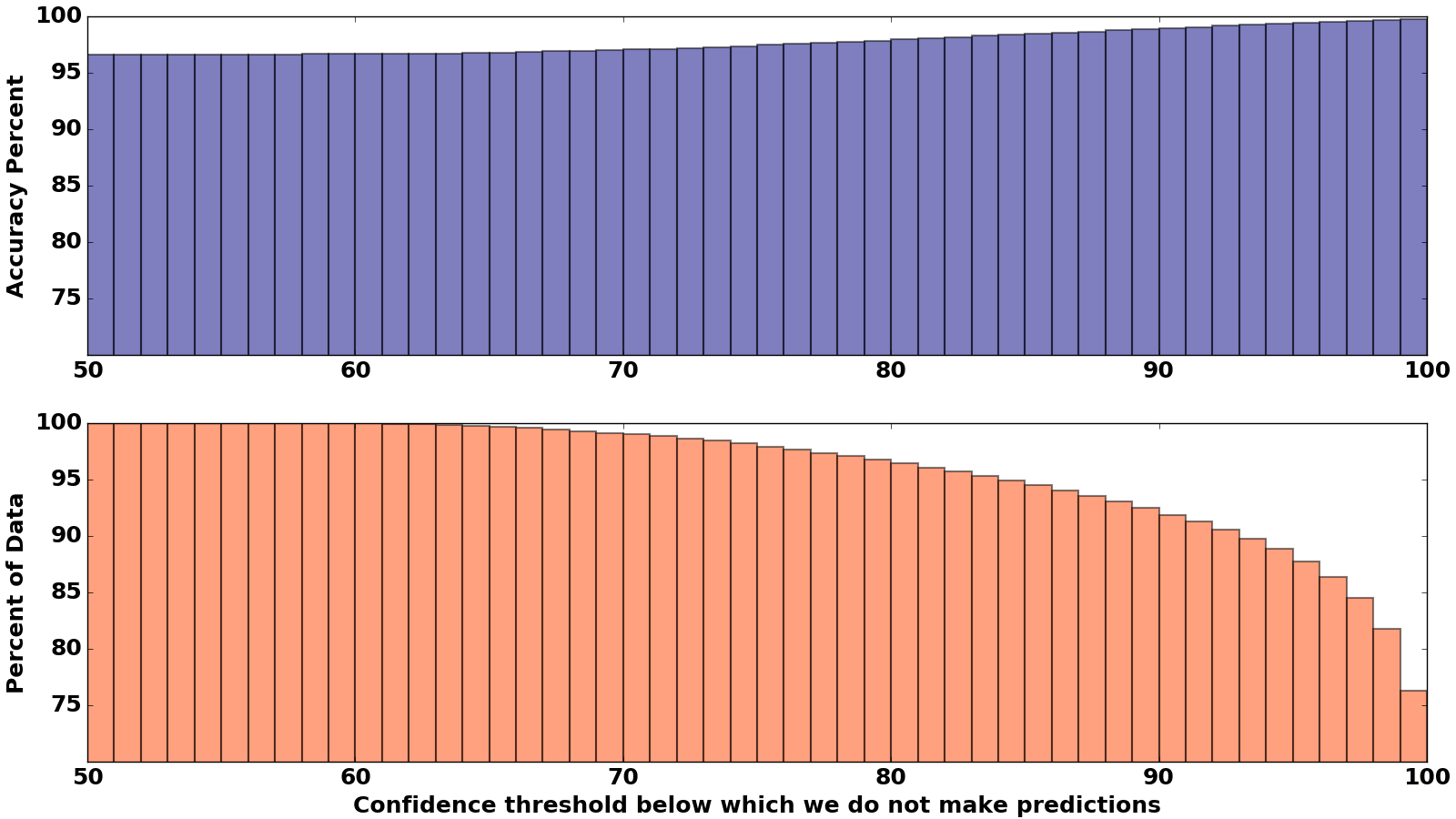}
	\caption{To increase the reliability of our model we can filter out the images that the network is not confident about and let experts label them instead.
Here we report the accuracy (top panel) of our VGG model on the images that are given confidence scores $\geq$  the thresholds (x-axis) for \nameref{sec:task1}	\textbf{Top:} The top-1 accuracy of the VGG model when we filter out images at different confidence levels (x-axis). \textbf{Bottom:} The percent of the dataset that remains when we filter out images for which that same model has low confidence. If we only keep the images that the model is 99\% or more confident about, then we can have a system with 99.8\% accuracy for 76\% of the data (rightmost column).}
	\label{fig:EFFilterOut} 
\end{figure}

\begin{figure}[h!]
	\setlength{\unitlength}{0.14in}
	\centering 
	\includegraphics[width=.5\textwidth]{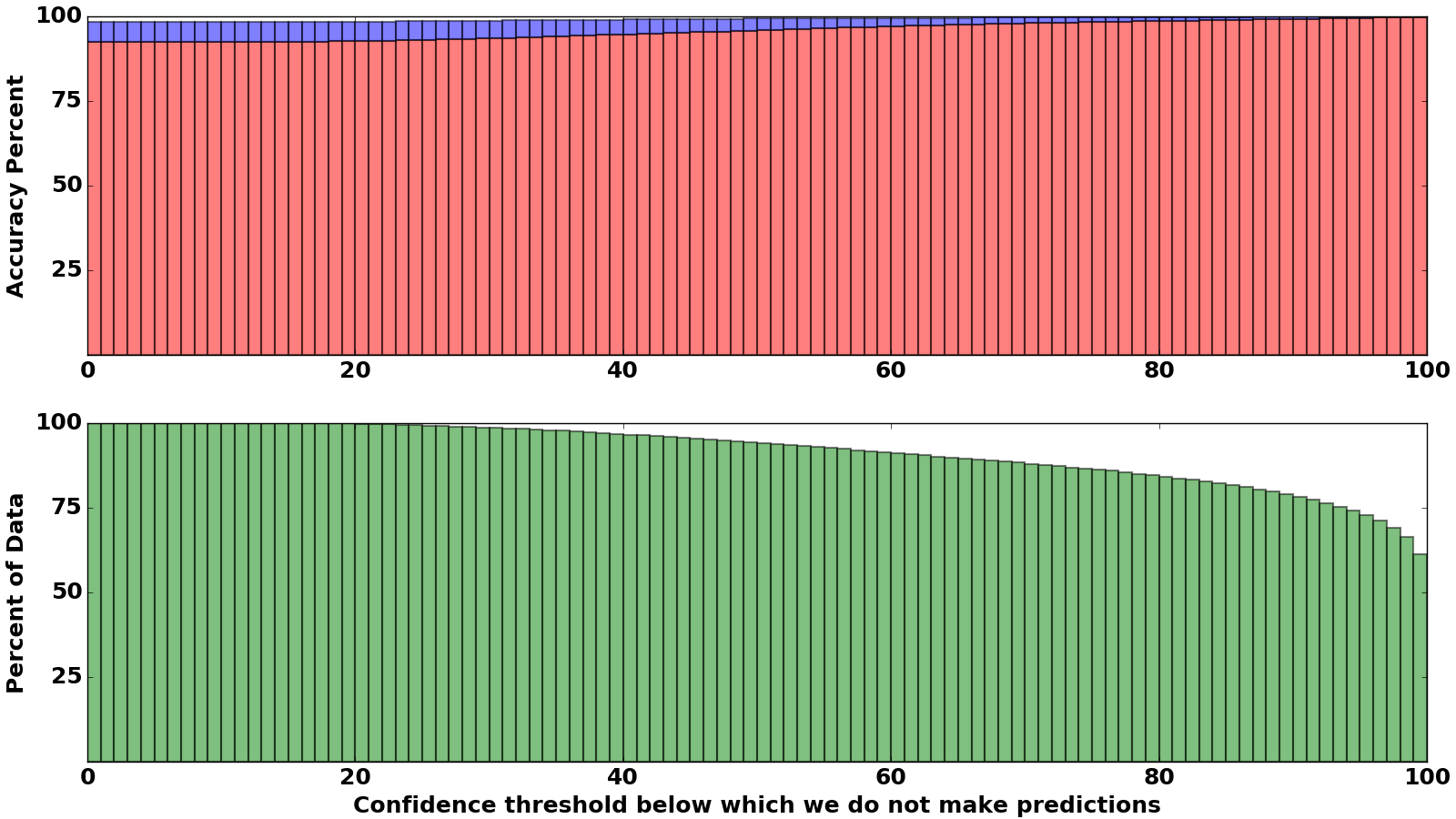}
	\caption{The figures are plotted in the same way as Fig.~\ref{fig:EFFilterOut}, but here for the ensemble of models for \nameref{sec:task2}.
	If we only keep the images that the model is 99\% or more confident about, we have a system that performs at 99.8\% top-1 accuracy on 66.1\% of the data (the rightmost column). \textbf{Top:} The top-1 (red) and top-5 (blue) accuracy of the ensemble of models when we filter out images with different confidence levels (x-axis).}
	\label{fig:IdentificationFilterOut} 
\end{figure}

\begin{figure}[h!]
	\setlength{\unitlength}{0.14in}
	\centering 
	\includegraphics[width=.5\textwidth]{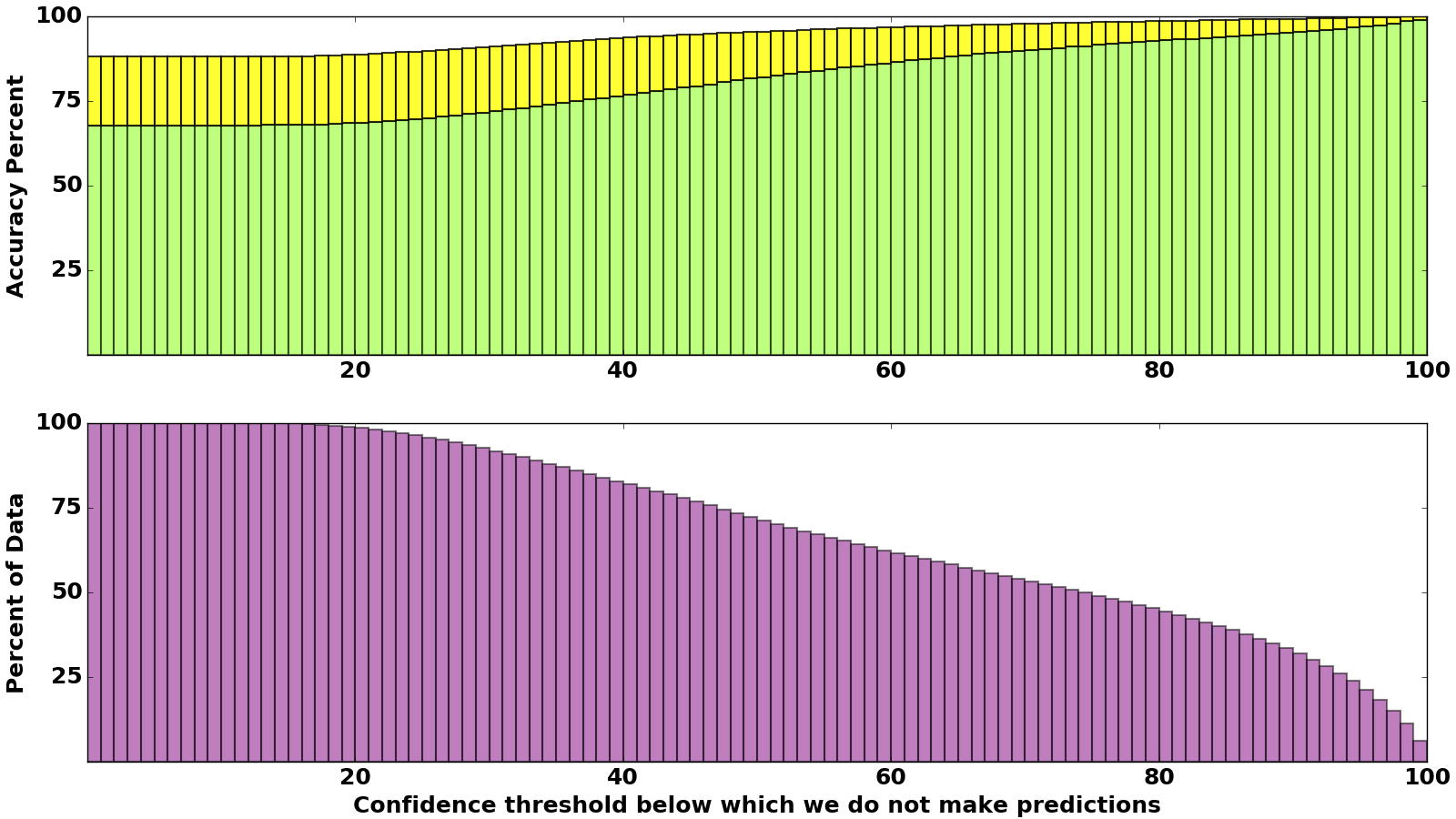}
	\caption{The figures are plotted in the same way as Fig.~\ref{fig:EFFilterOut} and Fig.~\ref{fig:IdentificationFilterOut}, but here for \nameref{sec:task3} and the ensemble of models. If we only keep the images that the model is 99\% or more confident about, we have a system that performs at 97.8\% top-1 accuracy on 8.4\% of the data (the rightmost column). \textbf{Top:} The top-1 (light green) and percent of predictions within +/- 1 bins (yellow) of the ensemble of models when we filter out images with different confidence levels (x-axis).}
	\label{fig:CountFilterOut} 
\end{figure}

The output probabilities per class (i.e. predictions) by deep neural networks can be interpreted as the confidence of the network in that prediction \cite{bridle1990probabilistic}. We can take advantage of these confidence measures to build a more accurate and more reliable system by automatically processing only those images that the networks are confident about and asking humans to label the rest. We threshold at different confidence levels, which results in the network classifying different amounts of data, and calculate the accuracy on that restricted dataset. We do so for \nameref{sec:task1} (Fig. \ref{fig:EFFilterOut}), \nameref{sec:task2} (Fig. \ref{fig:IdentificationFilterOut}), and \nameref{sec:task3} (Fig. \ref{fig:CountFilterOut}). As mentioned above, we cannot perform this exercise for \nameref{sec:task4} because SS lacks expert-provided labels for this task, meaning human-volunteer accuracy on it is unknown.

\section{Improving Accuracy for Rare Classes}
\label{sec:Imbalance}

\begin{figure*}[h]
	\setlength{\unitlength}{0.34in}
	\centering 
	\includegraphics[width=0.85\textwidth]{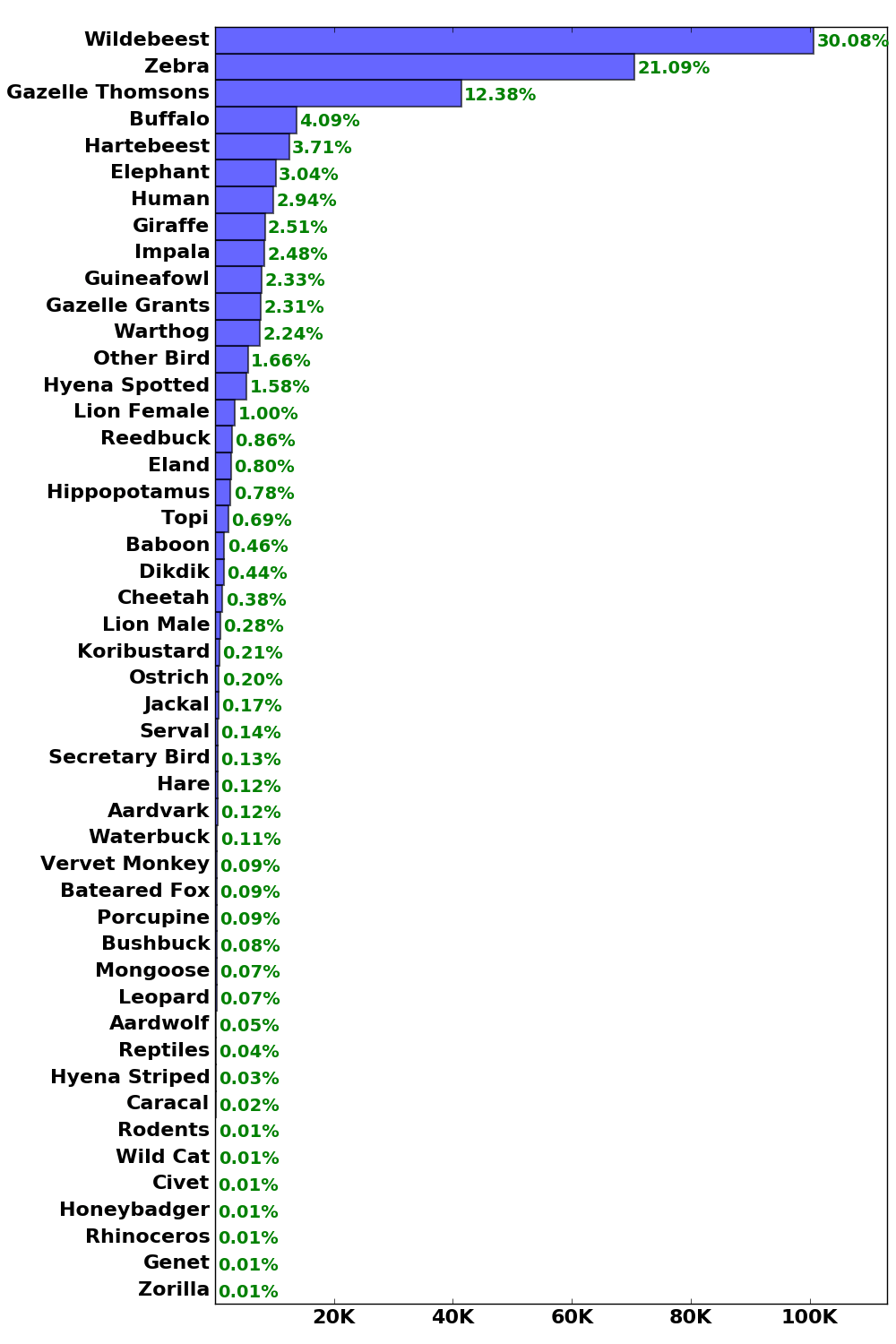}
	\caption{The number and percent of capture events belonging to each of the species. The dataset is heavily imbalanced. Wildebeests and zebras form $\sim$50\% of the dataset (top 2 bars), while more than 20 other species add up to only $\sim$1\% of the dataset (bottom 20 bars).}
	\label{fig:CaptureEvents} 
\end{figure*}

As previously mentioned, the SS dataset is heavily imbalanced. In other words, the numbers of available capture events (and thus pictures) for each species are very different (Fig.~\ref{fig:CaptureEvents}). For example, there are more than 100,000 wildebeest capture events, but only 17 zorilla capture events. In particular, 63\% of capture events contain wildebeests, zebras, and Thomson's gazelle. Imbalance can produce pathological machine learning models because they can limit their predictions to the most frequent classes and still achieve a high level of accuracy. For example, if our model just learns to classify wildebeests, zebras, and  Thomson's gazelle, still it can achieve 63\% accuracy while ignoring the remaining 94\% of classes. Experimental results show that our models obtain extremely low accuracy on rare classes (i.e. the classes with only few training examples) (Fig. \ref{fig:Imbalance}, bottom classes in the leftmost column have as low as $\sim$0\% accuracy scores). To ameliorate the problem caused by imbalance, we try three methods which we describe in the following subsections. All the following experiments are performed on the volunteer-labeled test set for the ResNet-152 model (which had the best top-1 accuracy on classifying all 48 SS species).

\subsection*{Weighted Loss}
For classification tasks, the measure of performance (i.e. accuracy) is defined as the proportion of examples that the models correctly classifies. In normal conditions, the cost associated with missing an example is equal for all classes. One method to deal with imbalance in the dataset is to put more cost on missing examples from rare classes and less cost for missing examples of the frequent classes, which we will refer to as the \emph{weighted loss} approach \cite{he2009learning}. For this approach, we have a weight for each class indicating the cost of missing examples from that class. To compute the weights, we divide the total number $N$ of examples in the set by the total number of examples $n_{i}$ from each class $i$ in the training set.  Then, we calculate the associated weights for each class using Eq. \ref{eq:fi} and \ref{eq:weights}. Because the dataset is highly imbalanced, we would have some very large class weights and some very small class weights for our method. Our models are trained by the backpropagation algorithm which computes the gradients over the network. These extreme weights result in very small or very large gradients, which can be harmful to the learning process. A quick remedy for this problem is to clamping the gradients within a certain range. In our experiments, we clamped the gradients of the output layer in the $[-0.01,0.01]$ range.

\begin{equation}
\label{eq:fi}
f_{i}=\frac{N}{n_{i}}
\end{equation}

\begin{equation}
\label{eq:weights}
w_{i}=\frac{f_{i}}{\sum_{i=1}^{48} f_{i}}
\end{equation}

The obtained results of this experiment (Fig. \ref{fig:Imbalance}, middle-left column) show that applying this method can increase the accuracy for the rare classes while keeping the same level of accuracy for most of the other classes. This method is especially beneficial for genet (40\% improvement) and aardwolf (35\% improvement). Applying the weighted loss method slightly hurts the top-1 accuracy, but it improved top-5 accuracy. The results suggest the weighted loss method is an effective way for dealing with imbalance in dataset.

\subsection*{Oversampling}
\begin{figure*}[h!]
	\centering 
	\includegraphics[width=0.9\textwidth]{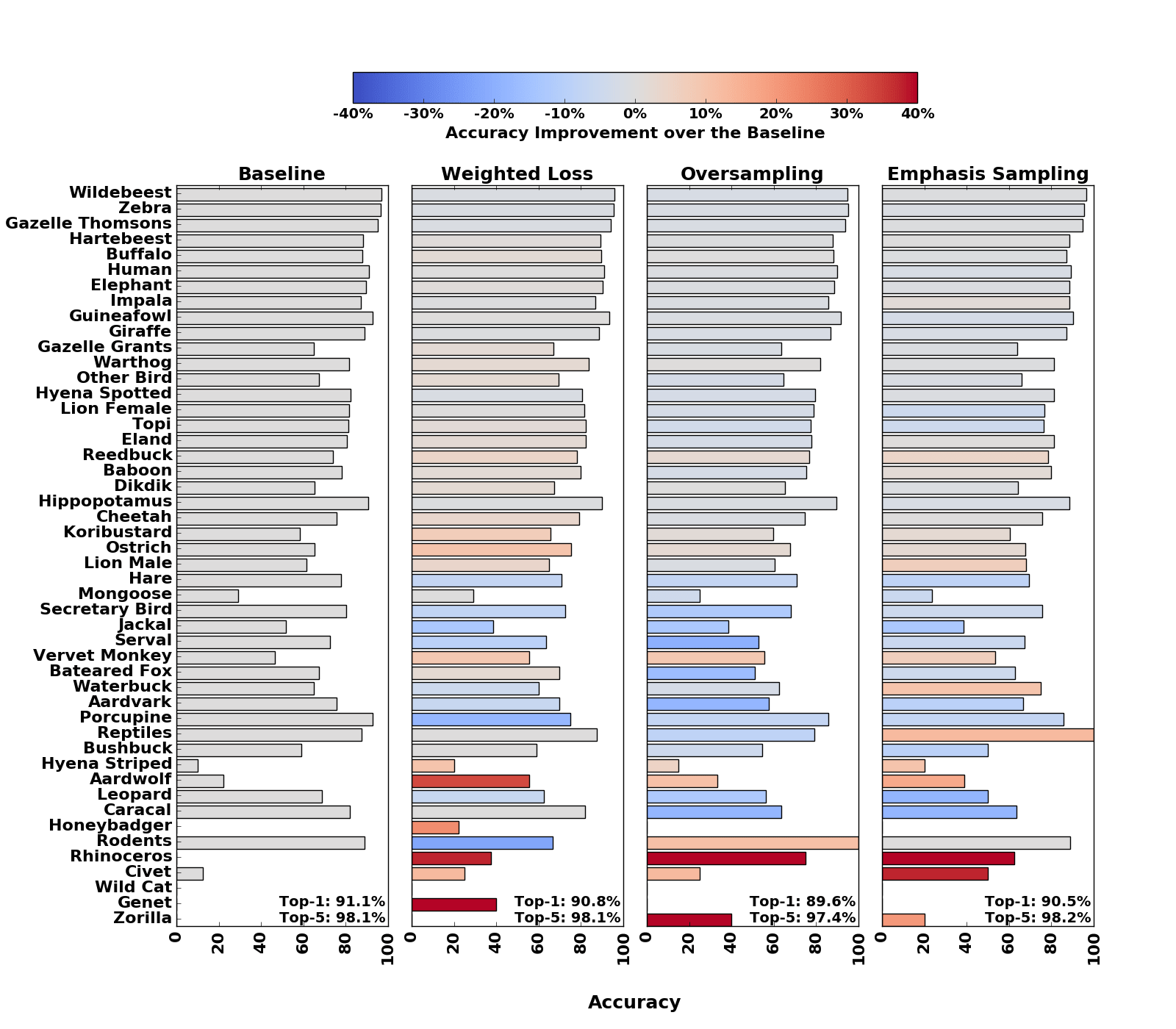}
	\caption{The effect of three different methods: weighted loss, oversampling, and emphasis sampling on the classification accuracy for each class. In all of them, the classification performance for some rare classes has been improved at the cost of losing some accuracy on the frequent classes. The color indicates the percent improvement each method provides. All three methods improved accuracy for several rare classes: for example, the accuracy for the rhinoceros class dramatically increases from near 0\% (original) to $\sim$40\% (weighted loss), $\sim$80\% (oversampling) and $\sim$60\% (emphasis sampling). Although the difference in global accuracies is not substantial, the weighted loss method has the best top-1 accuracy and the emphasis sampling method has the best top-5 accuracy. Moreover, it is notable that the emphasis sampling method has top-5 accuracy score of 98.2\% which is slightly higher than the 98.1\% accuracy of the baseline. In this plot, all classes are arranged based on their class sizes in descending order from the top to bottom.}
	\label{fig:Imbalance} 
\end{figure*}

Another method for dealing with dataset imbalance is \emph{oversampling} \cite{he2009learning}, which means feeding examples from rare classes more often to the model during training. This means that, for example, we show each sample in the zebra class only once to the model whereas we show the samples from the zorilla class around 4,300 times in order to make sure that the network sees an equal number of samples per class. The results from this experiment (Fig. \ref{fig:Imbalance}, middle-right column) show that the oversampling technique boosted the classification accuracy for rhinoceros ($\sim$80\%) and zorilla (40\%) classes.
We empirically found oversampling to slightly hurt the overall performance more than the other two methods (Fig.~\ref{fig:Imbalance}, the overall top-1 and top-5 accuracy are lower than those of the baseline, weighted loss and emphasis sampling methods). Further investigation is required to fully explain this phenomenon.

\subsection*{Emphasis Sampling}
Another method for solving the imbalance issue, which can be considered as an enhanced version of oversampling is \emph{emphasis sampling}. In emphasis sampling, we give another chance to the samples that the network fails on: the probability of samples being fed again to the network is increased whenever the network misclassifies them. Thus if the network frequently misclassifies the examples from rare classes it will be more likely to retrain on them repeatedly, allowing the model to make more changes to try to learn them.

For implementing the emphasis sampling method, we considered two queues, one for the examples that the top-1 guess of the network is not correct and one for the examples that all the top-5 guesses of the network are incorrect about. Whenever the model misclassifies an example, we put that example in the appropriate queue. During the training process, after feeding each batch of examples to the network, we feed another batch of examples taken from the front of the queues to the model with probability of 0.20 for the first queue and 0.35 for the second queue. Doing so, we increase the chance of wrongly classified images to be presented to the network more often. 

The results from this experiment (Fig. \ref{fig:Imbalance}, right-most column) indicate that this method can increase the accuracy for some of the rare classes, such civet ($\sim$40\%) and rhinoceros ($\sim$40\%). Moreover, emphasis sampling improved top-5 accuracy for the dataset in overall.  

\subsection*{Overall}
We found that all three methods perform similarly and can improve accuracy for some rare classes.
However, they do not improve the accuracy for \emph{all} the rare classes. More future research is required to further improve these methods.

\begin{figure*}
	\centering
	\subfloat[]{\includegraphics[width=0.30\textwidth]{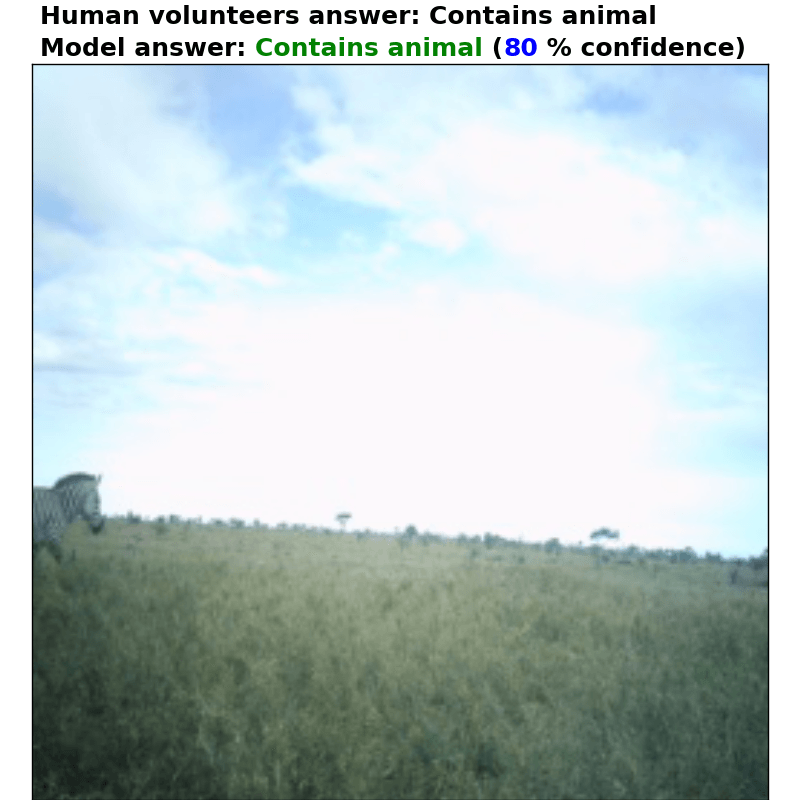} }%
	\subfloat[]{\includegraphics[width=0.30\textwidth]{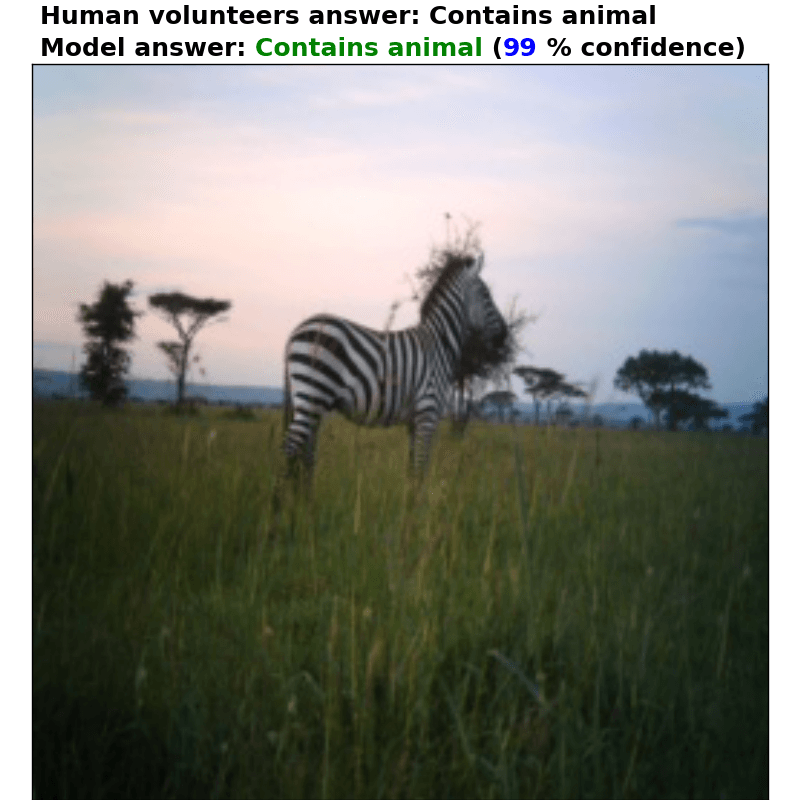} }%
	\subfloat[]{\includegraphics[width=0.30\textwidth]{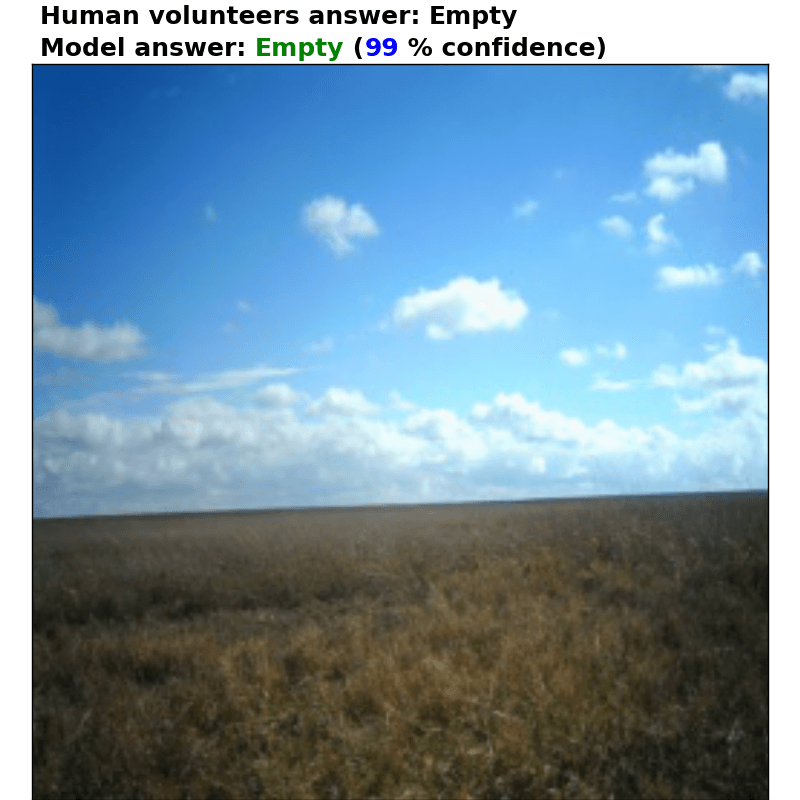} }%
	\hskip 5pt
	\subfloat[]{\includegraphics[width=0.30\textwidth]{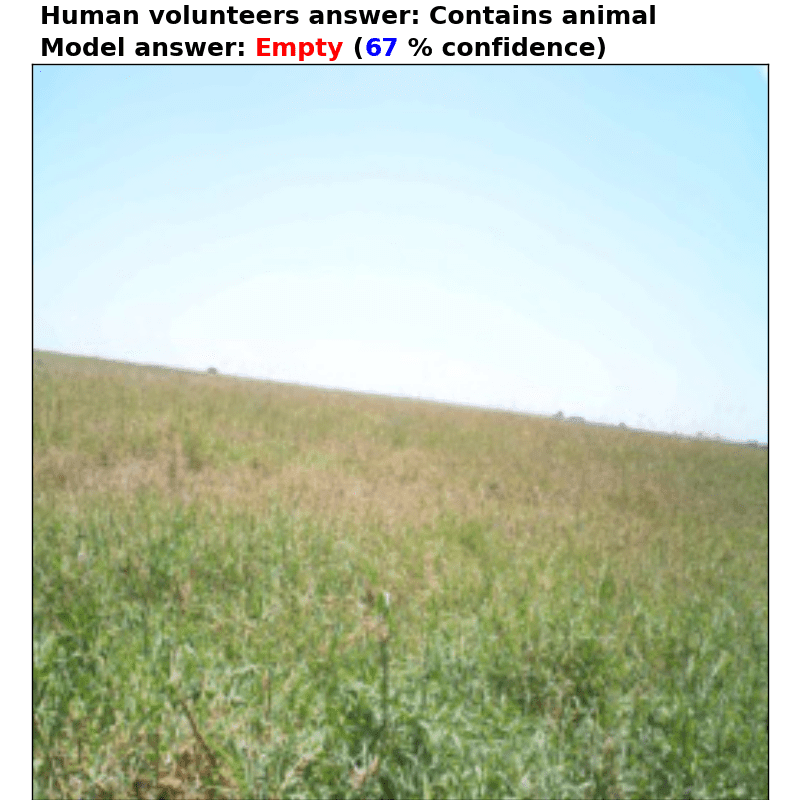} }%
	\subfloat[]{\includegraphics[width=0.30\textwidth]{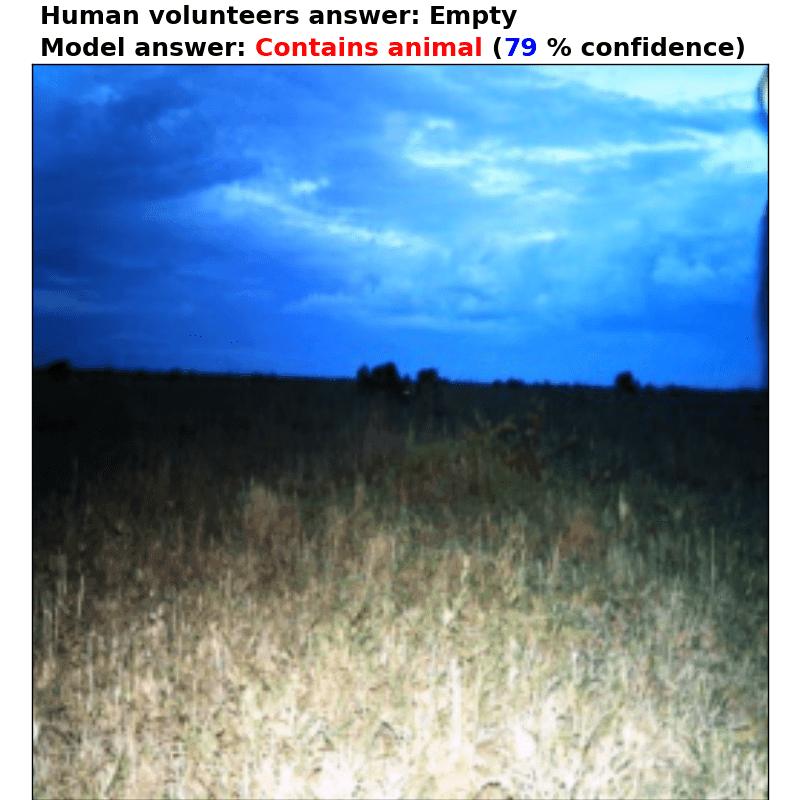} }%
	\subfloat[]{\includegraphics[width=0.30\textwidth]{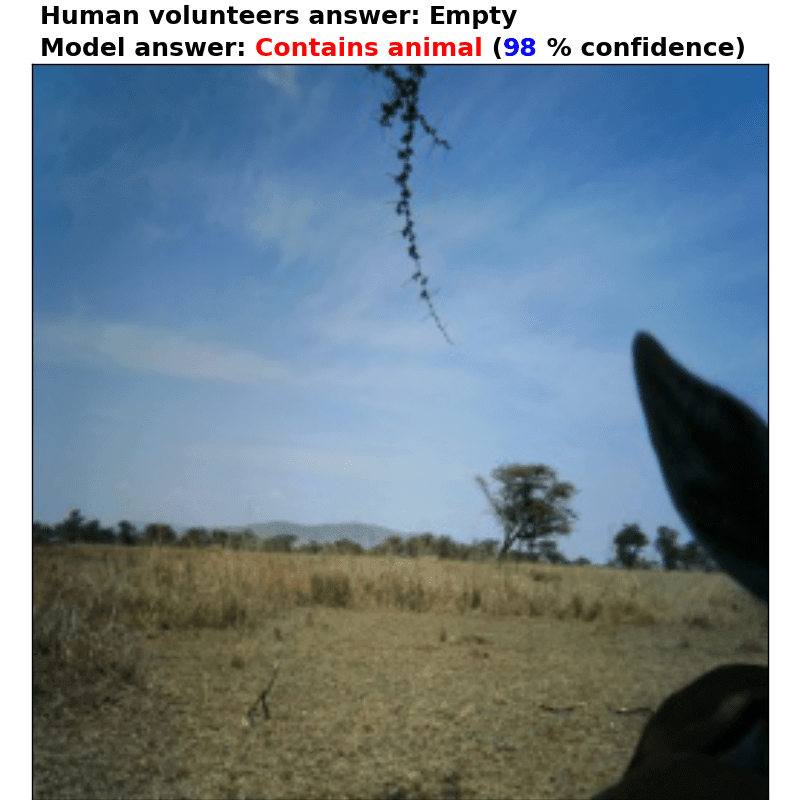} }%
	\hskip 5pt 
	\subfloat[]{\includegraphics[width=0.30\textwidth]{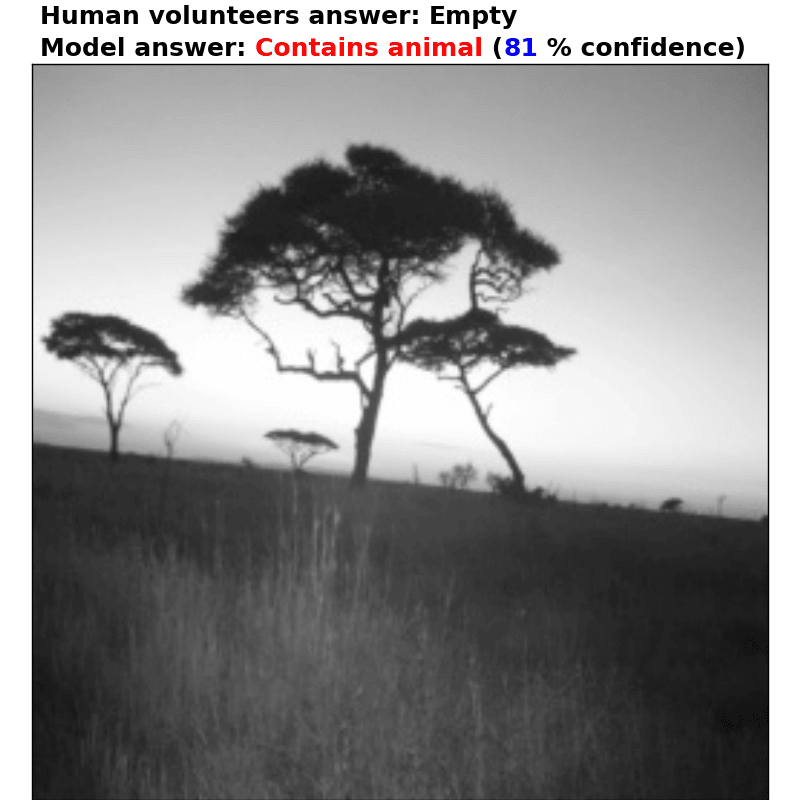} }%
	\subfloat[]{\includegraphics[width=0.30\textwidth]{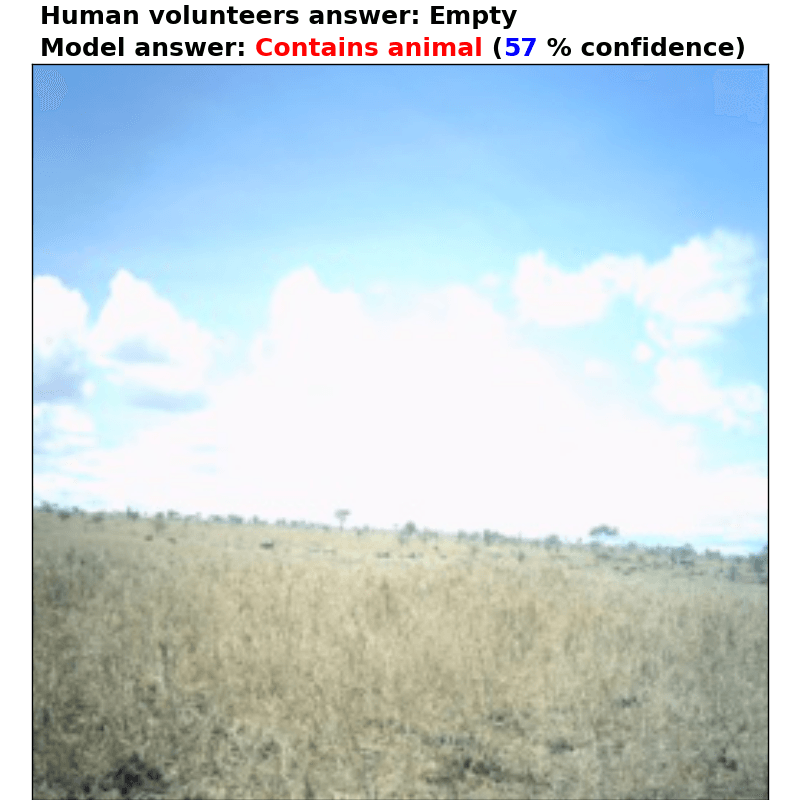} }%
	\subfloat[]{\includegraphics[width=0.30\textwidth]{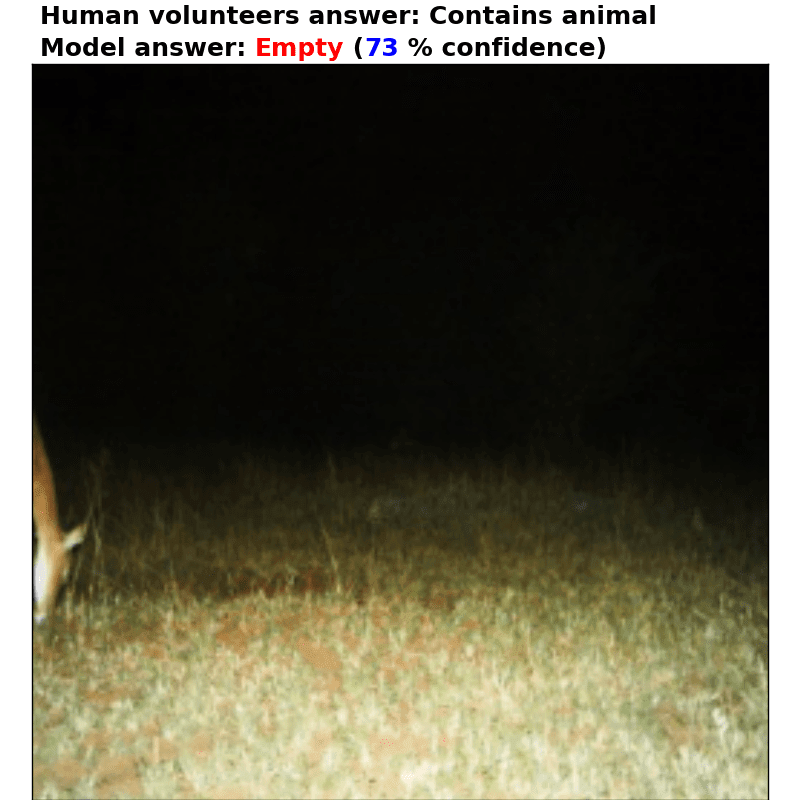} }%
	\caption{From the empty vs. animal task, shown are nine images, the human-volunteer answer, and the VGG network's answer along with its confidence. The first row of the images shows three correct answers by the model. The middle row shows three examples in which the model is correct, but volunteers are wrong, showing that volunteer labels are imperfect. The bottom row of images shows three examples in which volunteers are correct, but the model is wrong.
	}
	\label{fig:EFMistakes}
\end{figure*}

%
\end{document}